\newif\ifconfver
\confverfalse     %declaring conference version false
\newif\ifplainver  %declare a plain version
\plainvertrue
\plainverfalse

\ifplainver
\confverfalse   %automatically disable conf. version argument if it's plain
\fi

\ifconfver
\documentclass[10pt,twocolumn,twoside]{IEEEtran}
\else
\ifplainver
\documentclass[11pt]{article}
\usepackage{fullpage}
\else
\documentclass[11pt,draftcls,onecolumn]{IEEEtran}
\fi
\fi

\usepackage{xargs}
\usepackage{graphicx,algorithm,algorithmic,bm,amsmath,amsthm,amssymb,color,shortcuts_OPT,hyperref,cite,bbm,cleveref}
\usepackage[table]{xcolor}
\usepackage{varwidth}
\usepackage{subfigure}
\usepackage{caption}
\usepackage{slashbox}
\usepackage{booktabs}

\newtheorem{Example}{Example}

\newtheorem{assumption}{Assumption}

\definecolor{lavander}{cmyk}{0,0.48,0,0}
\definecolor{violet}{cmyk}{0.79,0.88,0,0}
\definecolor{burntorange}{cmyk}{0,0.52,1,0}

\definecolor{asuorange}{rgb}{0,0,1}
\definecolor{asured}{rgb}{0,0,1}
\definecolor{asuborder}{rgb}{0,0,1}
\definecolor{asugrey}{rgb}{0,0,1}
\definecolor{asublue}{rgb}{0,0,1}
\definecolor{asugold}{rgb}{0,0,1}

\newcommand\Ec{\ensuremath{\mathcal{E}}}

\newcommand\LLc{\ensuremath{{\mathcal{L}}}}

\newcommand\xb{\ensuremath{{\bm x}}}

\newcommand\Ab{\ensuremath{{\bm A}}}

\newcommand\Db{\ensuremath{{\bm D}}}

\newcommand\fb{\ensuremath{{\bm f}}}

\newcommand\gb{\ensuremath{{\bm g}}}
\newcommand\Ib{\ensuremath{{\bm I}}}
\newcommand\Lb{\ensuremath{{\bm L}}}

\newcommand\Sb{\ensuremath{{\bm S}}}

\newcommand\Yb{\ensuremath{{\bm Y}}}

\newcommand\Wb{\ensuremath{{\bm W}}}
\newcommand\hWb{\ensuremath{\hat{\bm W}}}

\newcommand\thetab{\ensuremath{{\bm \theta}}}

\newcommand\mub{\ensuremath{{\bm \mu}}}

\newcommand\Upsilonb{\ensuremath{{\bm \Upsilon}}}
\newcommand\zerob{\ensuremath{{\bm 0}}}
\newcommand\oneb{\ensuremath{{\bm 1}}}

\makeatletter
\def\multilimits@{\bgroup
	\Let@
	\restore@math@cr
	\default@tag
	\baselineskip\fontdimen10 \scriptfont\tw@
	\advance\baselineskip\fontdimen12 \scriptfont\tw@
	\lineskip\thr@@\fontdimen8 \scriptfont\thr@@
	\lineskiplimit\lineskip
	\vbox\bgroup\ialign\bgroup\hfil$\m@th\scriptstyle{##}$\hfil\crcr}
\def\Sb{_\multilimits@}
\def\endSb{\crcr\egroup\egroup\egroup}
\makeatother

\newtheoremstyle{t}         %name
{\baselineskip}{2\topsep}      %space above and below
{\rm}                   %Body font
{0pt}{\bfseries}  %Heading indent and font
{}                      %after heading
{ }                      %head after space
{\thmname{#1}\thmnumber{#2}.}

\theoremstyle{t}

\makeatletter
\DeclareRobustCommand*\cal{\@fontswitch\relax\mathcal}
\makeatother

\definecolor{violet}{cmyk}{0.79,0.88,0,0}
\usepackage[]{color-edits}
\addauthor{mh}{blue}
\addauthor{to}{blue}
\addauthor{thc}{blue}
\addauthor{st}{blue}

\begin{document}
	\title{Distributed Learning in the Non-Convex World: From Batch to Streaming Data, and Beyond}
	%\date{\today}
	%\vspace{-1cm}
	\author{Tsung-Hui Chang, Mingyi Hong,  Hoi-To Wai, Xinwei Zhang, Songtao Lu \thanks{THC, MH, HTW are ordered alphabetically, and contributed equally. MH is the corresponding author. THC is with the School of Science and Engineering, The Chinese University of Hong Kong, Shenzhen, China.
	MH and XZ are with the ECE Department, University of Minnesota, MN, USA.
	HTW is with the Department of SEEM, The Chinese University of Hong Kong, Hong Kong SAR, China. SL is with IBM Research AI, IBM Thomas J. Watson Research Center Yorktown Heights, New York 10598, USA. 
	
	THC (email: changtsunghui@cuhk.edu.cn) is supported in part by the National Key R\&D Program of China under Grant 2018YFB1800800, the NSFC, China, under Grant 61731018, and in part by the Shenzhen Fundamental Research Fund under Grant No. ZDSYS201707251409055 and No. KQTD2015033114415450.
	HTW (email:htwai@se.cuhk.edu.hk) is supported by the CUHK Direct Grant \#4055113, MH, SL and XZ (emails: \{mhong,zhan6234\}@umn.edu, songtao@ibm.com) are supported in part by NSF grants CMMI-172775, CIF-1910385 and by an ARO grant 73202-CS. }}
	\maketitle

\vspace{-1.5cm}
\begin{abstract}
%We are living in a highly connected world, which will become exponentially more connected soon.  The envisioned massive intelligence and connectedness is fueling a fundamental paradigm shift in information processing: The traditional centralized, server-client type models are going to be replaced by decentralized mechanisms at the network edge, which can effectively manage the {increasing} number of distributed intelligent devices,  {and} efficiently process the growing amount of data they generated.

Distributed learning has become a critical enabler of the massively connected world envisioned by many.
This article discusses four key elements {of} scalable distributed processing and real-time intelligence --- problems, data, communication and computation.
Our aim is to provide a fresh and unique perspective about how these elements should work together in an effective and coherent manner.
In particular, we {provide a selective review} about the recent techniques developed for optimizing {\it non-convex}  models {(i.e., problem classes)}, processing {\it batch and streaming} data {(i.e., data types)}, {over the networks} in a distributed manner {(i.e., communication and computation paradigm)}.
We describe the intuitions and connections behind {a core set of} popular distributed algorithms, emphasizing how to trade off between computation and communication costs.
{Practical issues and} future research directions will also be discussed.
%{We introduce a generic mathematical framework to model diverse data types and challenging non-convex models, and leverage it to develop intuitions and connections behind popular distributed algorithms,  and to discuss  future research directions.}
%\mhedit{to do: unify first-order, non-convex, step size}
\end{abstract}

\vspace{-0.2cm}
\section{Introduction}\label{sec:intro}

We are living in a highly connected world, and it will become exponentially more connected in a decade.  
By 2030, there will be more than 125 billion interconnected smart devices, creating a massive network of  intelligent appliances, cars, gadgets and tools (\url{https://developer.ibm.com/articles/se-iot-security/}).
% Soon there will be more than 50 billion smart devices interconnected to each other, creating a massive network of  intelligent appliances, cars, gadgets and tools \cite{cisco-iot}.
These devices collect a huge amount of real-time data, perform complex computational tasks, and provide vital services which significantly improve our lives and enrich our collective productivity. 

\begin{figure}[t]
	\begin{center}
		\includegraphics[width=0.43\linewidth]{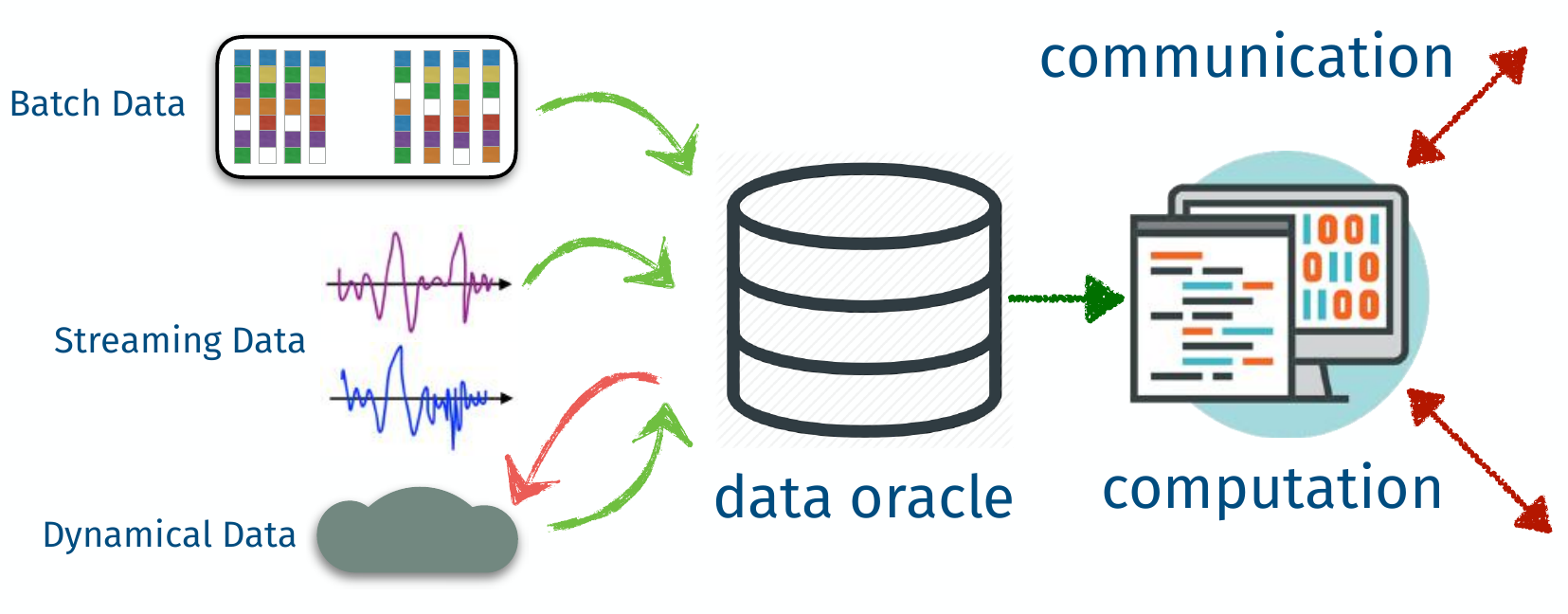}\hfill\includegraphics[width=0.54\linewidth]{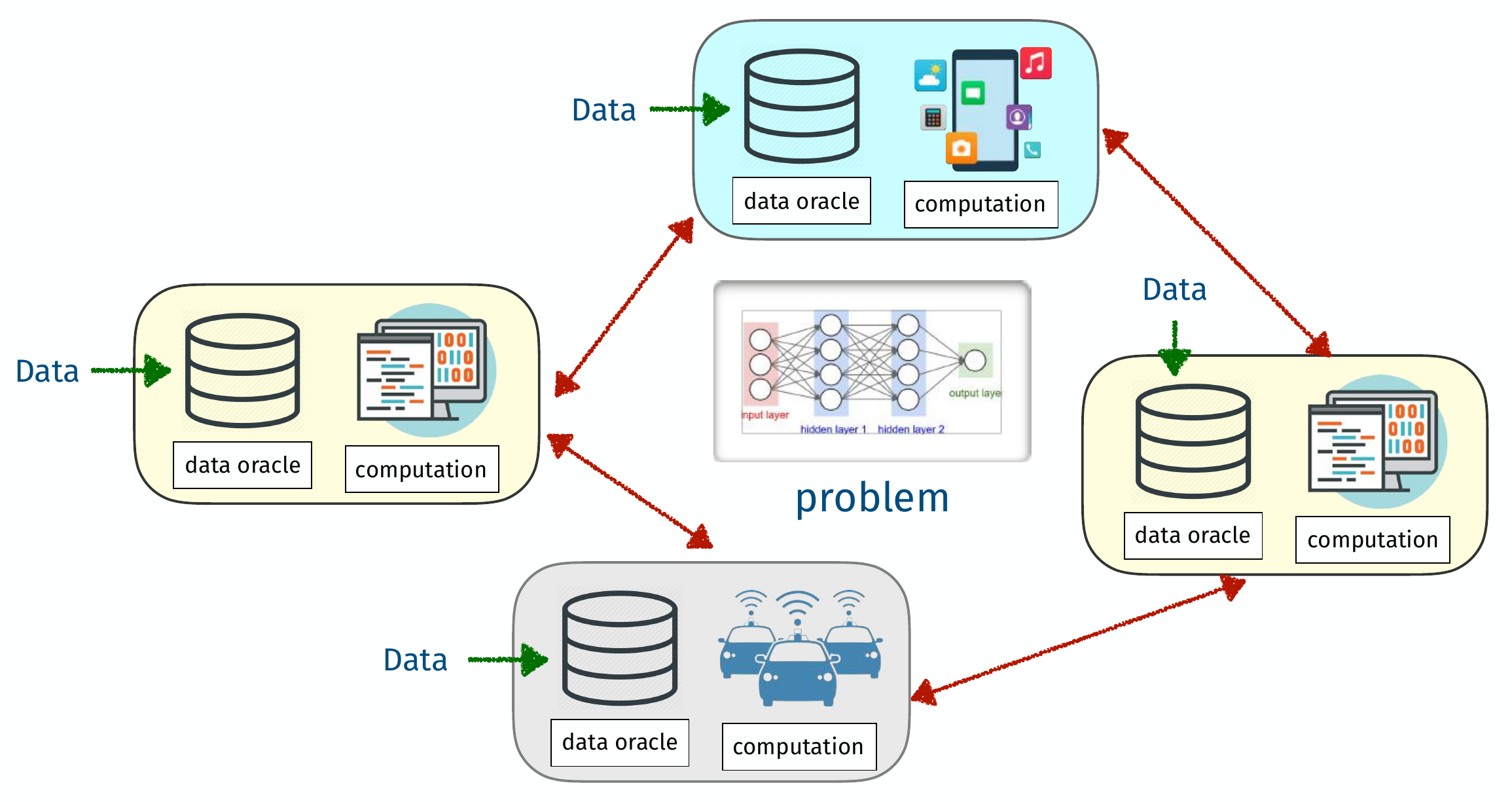}\vspace{-0.3cm}
		\caption{\footnotesize Overview of the key elements in distributed learning. (Left) Flow between different elements at a single agent: \textbf{data} (e.g., images) is taken from diverse types through an \emph{oracle},  \textbf{processed locally} before \textbf{communicating} with other agents, the final goal is to tackle a non-convex learning \textbf{problem}. (Right) Distributed learning on a network of agents.} \label{fig:overview}%\vspace{0.5cm}
	\end{center}
	\vspace{-1.5cm}
\end{figure}
There are four key elements {that enable} scalable distributed processing and real-time intelligence in a massively connected world --- problems, data, communication and computation. These elements are {closely  tied with} each other as illustrated in Fig.~\ref{fig:overview}. For example, without a meaningful machine learning (ML) problem, {crunching large amounts of} data using massive computational resources rarely lead to any actionable intelligence.  Similarly, despite all its sophisticated design and nice interpretation from neural sciences, modern neural networks may not be so successful without highly efficient computation methods. {The {\it overarching goal} of this {selective review} is to provide a fresh and unique perspective about how these elements should work together in the most effective and coherent manner, as to realize scalable processing, real-time intelligence, and ultimately contribute to the vision of a highly smart and connected world.}   Some key aspects of these elements to be taken into consideration are outlined below:  

{\bf Non-Convexity.}  For many emerging distributed applications,  the problems to be solved by distributed nodes will be highly {complicated}. For instance, in distributed ML, the nodes (e.g., mobile devices) jointly learn a model based on the local data (e.g., images on each device). To accurately represent the local data, the nodes are often required to use {\it non-convex} loss functions, 
such as the composition of multiple nonlinear activation functions in collaborative deep learning \cite{Wen2017,Daily2018,Jiang2017}. 

{\bf Data Acquisition Processes.} {One of the main reasons behind the recent success of ML is the ability to process data at scale. This not only means that one can process large volumes of data quickly (i.e., dealing with {\it batch} data), but more importantly, it requires  capability of dealing with {{{\it streaming} data.}} %and {\it dynamical} data. 
	There is an urgent need, and hence a growing  research, to deal with the massive amount of streaming data %\mhdelete{(\url{https://mashable.com/2013/09/16/facebook-photo-uploads/})}
	from online review platforms (e.g., Amazon), social network (e.g., Facebook), etc.
	%\thcdelete{Further beyond streaming data, in emerging ML paradigms such as reinforcement learning \cite{barto2018reinforcement}, the data (e.g., reward from playing the Go game) will be {\it dynamical} in the sense that the data generated depends on the algorithms designed to process the data itself. These types of data will complicate the inter-dependency between data, problems and computation, and present new challenges in algorithm design and analysis. }
	
	{\bf Distributed Processing.} {The growing} network size, the increased amount of the distributed  data, and the requirements for real-time response  often make traditional centralized processing not viable. 
	For example, self-driving cars should be carefully coordinated when meeting at an intersection, but since every such vehicle can generate up to 40 Gbit of data (e.g., from LIDAR and cameras) per second -- an amount that overwhelms the fastest cellular network -- it is impossible to pool the entirety of  data for real-time central coordination. This and other examples, from small and ordinary (e.g. coordinating smart appliance in home) to large and vitally important (e.g., national power distribution), show how paramount fast distributed processing will be to our collective well-being, productivity and prosperity. 
	
	{This paper is a {\it selective review} about recent advances on distributed algorithms. Unlike existing articles \cite{nedic2010coop,sayed2013diffusion,cevher2014convex}, this paper is centered around {\it non-convex}  optimization and learning problems. Our focus is to reveal connections and design insights about a core set of  first-order algorithms, while highlighting the interplay between problem, data, computation, and communication. % Importantly, we demonstrate how to choose the right algorithm for the learning task. \mhcomment{did we say this?}
		%Unlike existing articles \cite{nedic2010coop,sayed2013diffusion,cevher2014convex}, this paper is centered around {\it non-convex} learning problems}. %by first order methods and relying on {\it batch or streaming} data.} 
		We hope that the algorithm connections identified in this article will assist the readers in comparing theoretical and practical properties between algorithms, and will help translate new features and theoretical advances developed for one type of algorithms to other ``equivalent" types, without risking to ``reinventing the wheel".} %More importantly,  once additional features are developed for one type of algorithm, it is possible to reuse most of the existing theoretical and empirical properties without risking to ``reinventing the wheel".}
} %by first order methods and relying on {\it batch or streaming} data.

%\mhcomment{deleted 3 survey, can add it back somewhere.} 

We will start with a generic model of distributed optimization, taking into consideration diverse data types and non-convex models (Sec.~\ref{sec:framework}). We will then review state-of-the-art algorithms that deal with batch/streaming data, {{make useful connections between them}}
%in the distributed non-convex setting 
(Sec.~\ref{sec:algorithms}), and discuss practical issues in implementing these algorithms (Sec.~\ref{sec:practical}).  Finally, we discuss future research directions (Sec.~\ref{sub:discussion}). %\mhedit{We emphasize that the goal of this article is not to provide an extensive list of all relevant algorithms. Rather, our focus is given to a subset of relatively popular methods, and discuss the connections behind them. }\vspace{-.3cm}

\section{Problems and Data Models}\label{sec:framework}
%\vspace{-0.3cm}
%In this section, we describe the problem and data models for distributed learning. Moreover, we emphasize on the challenges in handling non-convex optimization in a distributed setting.\vspace{-.2cm}

%\subsection{Problem Class}
%\vspace{-0.3cm}
\textbf{Problem Class}. Consider $n$ inter-connected agents. The network connecting these agents is represented by a (directed or undirected) graph $G = ({\cal V}, {\cal E})$, where ${\cal V} = \{1,...,n\}$ is the set of agents and ${\cal E} \subseteq {\cal V} \times {\cal V}$ is the set of communication links between agents. The goal for agents is to find a solution $\param^\star := (\prm_1^\star, ..., \prm_n^\star)$ which tackles the \emph{non-convex} optimization problem: 
\beq \label{eq:opt}
\begin{array}{rl}
	\ds \min_{ \prm_i \in \RR^d,\forall\!\!~i } & \ds \frac{1}{n} \sum_{i=1}^n f_i ( \prm_i )~~\text{s.t.}~~
	(\prm_1, ..., \prm_n) \in \Param := \big\{ \prm_i \in \RR^d,~i=1,...,n : \prm_i = \prm_j,~\forall~(i,j) \in {\cal E} \big\}.
	%\prm_i \in \Prm,~\prm_i = \prm_j,~\forall~(i,j) \in {\cal E},
\end{array}
\eeq
%\mhcomment{this assumption is not correct, the entire function f is in this domain.}
{We define $f(\param) := \frac{1}{n} \sum_{i=1}^n f_i ( \prm_i )$}, where {$f : \Param \rightarrow \RR \cup \{ \infty \}$} is a (possibly non-convex) cost function of the $i$th agent.
% An important case which we will focus on is 
{Problem \eqref{eq:opt} contains a {coupling constraint} that enforces \emph{consensus}.} 
When $G$ is undirected and connected, we have $\prm_1 = \cdots = \prm_n$, 
and an optimal solution to \eqref{eq:opt} is a minimizer  to the following {\it equivalent} optimization problem 
$$\min_{\prm\in\mathbb{R}^d}\quad \frac{1}{n}\sum_{i=1}^n f_i(\prm).$$ 
{The consensus formulation
	motivates a {\it decentralized} approach to finding high-quality solutions, in the sense
	that each agent $i$ can only access its local cost function $f_i$ and process its local data together with messages
	exchanged from its neighbors. %Problem \eqref{eq:opt} has wide applications in signal
	%processing and ML, such as the collaborative
	%learning problem mentioned in Sec. \ref{sec:intro}.
}
%Problem \eqref{eq:opt} is a parameter estimation problem which finds the best fit $\prm$ to the available data, and it also has wide applications in signal processing and ML; see Section~\ref{sec:data}. 

Without assuming convexity for \eqref{eq:opt}, one cannot hope to find an optimal solution using a reasonable amount of effort, as solving a non-convex problem is in general NP-hard \cite{Murty1987}. Instead, %of seeking a globally optimal solution to \eqref{eq:opt}, 
we resort to finding \emph{stationary, consensual solutions} whose gradients are small and the variables are in consensus. Formally, let $\epsilon \geq 0$, we say that $\param = (\prm_1,...,\prm_n)$ is an $\epsilon$-stationary solution to \eqref{eq:opt} if 
\begin{align}\label{eq:stationarity} \textstyle
{\sf Gap}(\param) \eqdef \big\| n^{-1} \sum_{j=1}^{n}\grd f_j \big( \bar{\prm} \big) \big\|^2 + \sum_{j=1}^{n}\|\prm_j-\bar{\prm}\|^2 \leq \epsilon, \;\; \mbox{where}\;\;  \bar{\prm}:=n^{-1} \sum_{i=1}^n \prm_i.
\end{align} 
%Clearly, when $\epsilon=0$, $\prm_i$'s are consensual and they become a first-order stationary solutions of \eqref{eq:opt}.  
Below we summarize two commonly used conditions when approaching problem \eqref{eq:opt}.
\vspace{-0.2cm}
\begin{assumption}\label{ass:gen}
	{\sf (a)} The graph $G$ is undirected and connected. {\sf (b)} For $i=1,...,n$, the cost function $f_i(\prm)$ is $L$-smooth, satisfying the following condition: \vspace{-.1cm}
	\beq \label{eq:smooth}
	\| \grd f_i ( \prm ) - \grd f_i (\prm' ) \| \leq L \| \prm - \prm' \|,~\forall~\prm,\prm' \in \RR^d.
	\eeq
	Further, {$f : \Param \rightarrow \RR \cup \{ \infty \}$, that is, the average function $f$ is lower bounded over the domain $\Param$.}
	\vspace{-0.3cm}
\end{assumption}
Assumption~\ref{ass:gen}, together with \eqref{eq:opt}, describes a general setup for distributed learning problems. Our goal is to find a stationary, consensual solution satisfying \eqref{eq:stationarity}. 
{We remark that several recent works \cite{hong2018gradient,daneshmand2018second-arxiv,Swenson19annealing,vlaski2019distributed,vlaski2019distributedb} have analyzed the more powerful forms of convergence, such as to second-order stationary solutions or global optimal solutions. However, establishing these results requires additional assumptions, which further restrict the problem class, so we shall omit discussing these works in  detail due to space limitation.}

Having fixed the problem class, a distributed learning system consists of a {data acquisition} and {local processing} step performed at a local agent, and a {communication} step to exchange information between  agents. We summarize these key elements and their interactions in Fig.~\ref{fig:overview}. 
%We discuss the {data} acquisition models.  

%\mhcomment{Need to define what we meant by``distributed algorithms".}

% \vspace{-0.2cm}
%\subsection{Batch and Streaming Data} \label{sec:data}
\textbf{Data Model}.  {We adopt a {\it local oracle model [denoted as {${\sf DO}_i ( \prm_i )$}}] to describe how information about the {cost function} $f_i$ is retrieved in distributed learning. As we shall focus on first-order algorithms in the sequel, the oracle ${\sf DO}_i (\prm_i)$ is characterized as various estimates of the gradient $\nabla f_i(\prm_i)$.} 
%The following models are studied:

\paragraph{Batch Data} This is a classical setting where the entire local data set is available at \emph{anytime}, also known as the \emph{offline} learning setting. Denote $\State_{i,1}, \State_{i,2},..., \State_{i,M_i}$ as the local dataset of agent $i$ with $M_i$ data samples. The local cost function and DO are given by the \emph{finite sums} below:
\beq \textstyle \label{eq:emp}
f_i( \prm_i) = M_i^{-1} \sum_{\ell=1}^{M_i} F_i( \prm_i; \State_{i,\ell} ),~~{\sf DO}_i ( \prm_i^{t} ) = \grd f_i( \prm_i^{t} ) %= M_i^{-1} \sum_{\ell=1}^{M_i} \grd_{\prm} F_i( \prm_i^{(t)}; \State_{i,\ell} )
\eeq
where $F_i( \prm_i; \State_{i,\ell} )$ is the cost function corresponding to the $(i,\ell)$th data.

\paragraph{Streaming Data} In this setting, the data are revealed in a \emph{streaming} (or \emph{online}) fashion. We first specify our cost function as a stochastic function $f_i( \prm_i ) = \EE_{ \xi_i \sim \pi_i (\bm{\cdot}) } \big[ F_i ( \prm_i ; \xi_i ) \big]$, 
where $\pi_i( \bm{\cdot} )$ is a probability distribution of $\xi_i$.  
At each iteration {$t$}, querying the DO draws $m_t$ independent and identically distributed (i.i.d.) samples for the learning task; thus,
\beq \label{eq:stream} 
{\sf DO}_i ( \prm_i^{t} ) = m_t^{-1} {\textstyle \sum_{\ell=1}^{m_t} } \grd F_i( \prm_i^{t} ; \State_{i,\ell}^{t+1} )~~\text{where}~~\State_{i,\ell}^{t+1} \sim \pi_i (\bm{\cdot} ),~\ell=1,...,m_t,
\eeq 
which is an unbiased estimate of gradient, i.e., $\EE_{ \xi \sim \pi_i( \bm{\cdot})}[ {\sf DO}_i ( \prm ) ] = \grd f_i ( \prm )$. Moreover,
\begin{assumption} \label{ass:variance}
	Consider the DO \eqref{eq:stream} and random samples $\State$ drawn i.i.d.~from $\pi_i (\bm{\cdot} )$. Assume
	\beq \label{eq:var}
	\EE_{ \xi \sim \pi_i( \bm{\cdot})}[ \| {\sf DO}_i ( \prm ) - \grd f_i ( \prm ) \|^2 ] \leq {\sigma} %^2 
	< \infty,~i=1,...,n,~\forall~\prm \in \RR^d.
	\eeq
	%for $i=1,...,n$ and any $\prm \in \RR^d$.
\end{assumption}
\vspace{-0.3cm}
%Note that the first condition in \eqref{eq:var} is satisfied by \eqref{eq:stream} as we draw i.i.d.~samples. For the second condition, 
%we require $\grd F_i$ to be bounded and
In other words, the random variable ${\sf DO}_i ( \prm )$ has a bounded variance.
%The above assumptions are commonly found in the stochastic gradient literature.
%In the centralized learning setting, the DO (with $m_t = 1$) is equivalent to the stochastic gradient used in stochastic gradient (SGD) method \cite{}.

A related setting involves a large but fixed dataset ($M_i \gg 1$) at agent $i$, denoted by $\{\State_{i,1}, ..., \State_{i,M_i} \}$ and $f_i$ is given by \eqref{eq:emp}. Accessing the full dataset  entails an undesirable ${\cal O}(M_i)$ \emph{computation} complexity. As a remedy, we can draw at each iteration a small batch of random samples ($m \ll M_i$) \emph{uniformly} from the large dataset. This results in a DO akin to \eqref{eq:stream}. 
%While the DO remains unbiased with $\EE[ {\sf DO}_i ( \prm_i^{t} ) ] = \grd f_i (\prm_i^{t})$, the variance is bounded by $\max_{k,\ell} \|  
%\grd F_i( \prm_i^{t} ; \State_{i,k} ) - \grd F_i( \prm_i^{t} ; \State_{i,\ell} ) \|^2$.
%, where the bound quantifies the imbalance between the local datasets.

\vspace{-0.3cm}
{\subsection{Examples and Challenges}\label{sub:challenge}\vspace{-.3cm}
	We conclude this section by listing a few popular examples of non-convex learning problems and how they fit into the described models above. Moreover, we discuss the challenges with non-convex distributed learning, which motivate many algorithms reviewed in this paper.}

\vspace{-0.3cm}
\begin{Example}\label{exp:class}
	\emph{(Binary Classifier Training with Neural Network)} For each $i \in \{1,...,n\}$, suppose that a stream of training data $ \State_{i,1}^{t},\State_{i,2}^{t},...$ is available at the $i$th agent, where $\State_{i,j}^{t} = ( {\bm x}_{i,j}^{t}, y_{i,j}^{t} )$ is a tuple containing the feature ${\bm x}_{i,j}^{t} \in \RR^m$ and label $y_{i,j}^{t} \in \{ 0, 1 \}$. Let $\prm = ( {\bf W}^{(1)}, ..., {\bf W}^{(L)} )$ be the parameters of an $L$-layer neural network, we consider the models \eqref{eq:opt} with the following logistic loss:
	\beq \label{eq:logist loss}
	%\begin{split}
	%& 
	\textstyle F_{i} (\prm; \xi_{i,j}^{t} ) = (1-y_{i,j}^{t}) \log (1-h_{\prm}({\bm x}_{i,j}^{t}) ) + y_{i,j}^{t} \log h_{\prm} ({\bm x}_{i,j}^{t} ) ,% \\
	%& \text{where}~~h_{\prm}({\bm x}_{i,\ell}) = 
	%\end{split}
	\eeq
	where $h_{\prm}({\bm x}_{i,j}^{t})$ is the sigmoid function $(1+ g( {\bm x}_{i,j}^{t} ; \prm) )^{-1}$ such that $g( {\bm x}_{i,j}^{t} ; \prm)$ is the soft-max output of the last layer of the neural network with $ {\bm x}_{i,j}^{t}$ being the input. The hidden layer of the neural network may be defined as ${\bm g}^{(\ell+1)} = u( {\bf W}^{(\ell+1)} {\bm g}^{(\ell)} )$ for $\ell=0,...,L-1$, where $u(\cdot)$ is an activation function and ${\bm g}^{(0)} = {\bm x}_{i,j}^{t}$. The goal of \eqref{eq:opt} is to find a set of optimal parameters of a neural network, taking into account the (potentially heterogeneous) data received at {all agents}.
	Here, the loss function $f_i(\prm)$ is non-convex but satisfies \textbf{Assumption~\ref{ass:gen}}, and the DO follows the \textbf{streaming data} model.
\end{Example}
\vspace{-0.3cm}
\begin{Example}\label{exp:mat}
	\emph{(Matrix Factorization)}  The $i$th agent has a fixed set of $M_i$ samples where the $\ell$th sample is denoted as $\xi_{i,\ell} = {\bm x}_{i,\ell} \in \RR^{m_1}$. The data received at the agents can be encoded using the columns of a \emph{dictionary matrix} $\bm{\Phi} \in \RR^{m_1 \times m_2}$, \ie ${\bm x}_{i,\ell} \approx \bm{\Phi} {\bm y}_{i,\ell}$. The goal is to learn a factorization with the \emph{dictionary} $\bm{\Phi}$ and codes ${\bm Y}_i = ( {\bm y}_{i,1}~\cdots~ {\bm y}_{i,M_i} )$. Let ${\bm X}_i = ( {\bm x}_{i,1}~\cdots~ {\bm x}_{i,M_i} )$ be the data. The learning problem is: 
	\beq \label{eq:dictlearn}
	\min_{ \bm{\Phi}, {\bm Y}_i, i=1,...,n }~\frac{1}{n} \sum_{i=1}^n \left\| {\bm X}_i - \bm{\Phi} {\bm Y}_i \right\|_{\rm F}^2~~{\rm s.t.}~~\bm{\Phi} \in {\sf A},~{\bm Y}_i \in {\sf Y}_i,~i=1,...,n,
	\eeq
	%where ${\sf A}$, and ${\sf Y}_i$ represent some constraints on the dictionary and codes to ensure identifiability. An interesting aspect of \eqref{eq:dictlearn} is that the problem involves a \emph{common} variable $\bm{\Phi}$ and a \emph{private} variable ${\bm Y}_i$. The common variable has to be jointly decided by the data received at agents, while the private variables are nuisance parameters decided locally. 
	where ${\sf A}, {\sf Y}_i$ represent some constraints on the dictionary and codes to ensure identifiability.
	An interesting aspect of \eqref{eq:dictlearn} is that the problem optimizes a \emph{common} variable $\bm{\Phi}$ and a \emph{private} variable ${\bm Y}_i${; in particular, the corresponding local cost $f_i(\bm{\Phi})=\min_{\Yb_i \in {\sf Y}_i} \| {\bm X}_i - \bm{\Phi} {\bm Y}_i \|_{\rm F}^2$.} The common variable is jointly decided by the data received at agents, while the private variables are nuisance parameters decided locally. 
	Here, the loss function $f_i(\prm)$ satisfies \textbf{Assumption~\ref{ass:gen}}, and the DO follows the  \textbf{batch data} model.\vspace{-0.2cm}
\end{Example}
{Handling non-convex distributed learning problems involves several unique challenges.}
First, {directly applying algorithms developed for convex problems to the non-convex setting may lead to unexpected algorithm behaviors.} %\mhdelete{despite the fact that there are a plethora of algorithms developed for (centralized)  convex optimization,  it is not straightforward to one may be tempted} \mhdelete{While this may work out-of-the-box in some cases, a naive application may lead to unexpected algorithm behaviors. one cannot \emph{directly} apply an algorithm designed for {convex problems} and hope to find a good solution for the non-convex problem \eqref{eq:opt}, \eqref{eq:consensus}.} %This highlights the necessity to adopt an algorithm that involves a \emph{co-design} of \textbf{communication} and \textbf{computation} protocols. 
To see this, consider a simple example as follows. 
\vspace{-0.2cm}
\begin{Example}\label{exp:diverge}
	Consider \eqref{eq:opt} with $d=1$, $n=2$ agents connected via one edge.  {{Let $f_1(\theta_1 ) = \theta_1^2 / 2$, and $f_2(\theta_2) = -\theta_2^2 / 2$,}} where $f_2$ is \emph{non-convex}.
	{{Note that any $\theta_1=\theta_2$  is an optimal solution to the consensus problem}.}
	%This problem is equivalent to a  {\it consensus problem} since any $\theta_1=\theta_2$  is an optimal solution. 
	However, applying the classical distributed gradient descent (DGD) method \cite{Nedic09subgradient} [to be discussed in Sec. \ref{sub:batch}; see \eqref{eqn: DGD}], with a {\bf constant step size} $\gamma > 0$, generates the following iterates: %$\gamma$ and the mixing matrix as ${\bm W} = \frac{1}{2} {\bf 1}{\bf 1}^\top$ result in the following recursion:
	\beq \label{eq:diverge}
	\prm^{t+1}: = \left( \begin{array}{c}
		\theta_1^{t+1} \\ \theta_2^{t+1}
	\end{array} \right) = \left( \begin{array}{cc}
		\frac{1}{2} & \frac{1}{2} \\
		\frac{1}{2} & \frac{1}{2}
	\end{array} \right) \prm^{t} - \gamma \left( \begin{array}{c}
		\theta_1^{t} \\ -\theta_2^{t}
	\end{array} \right)
	= \underbrace{\left( \begin{array}{cc}
			\frac{1}{2} - \gamma & \frac{1}{2} \\
			\frac{1}{2} & \frac{1}{2} + \gamma 
		\end{array} \right)}_{\eqdef {\bm M}(\gamma)} \prm^{t} %\left( \begin{array}{c}
	%\theta_1^{(t)} \\ \theta_2^{(t)}
	%\end{array} \right) %= 
	%{\bm A}(\gamma) \prm^{(t)}.
	\eeq
	For all $\gamma > 0$, the spectral radius of ${\bm M}(\gamma)$ is larger than one, so  the above iteration always diverges. On the contrary, it can be verified that DGD converges linearly to a solution satisfying $\theta_1=\theta_2$ with {\it any} positive step sizes if we change the objective functions to {{$f_1(\theta_1)=f_2(\theta_2)=0$.} {Generally, if the problem is convex, DGD converges to a neighborhood of the optimal solution when small constant step sizes are used. This is in contrast to \eqref{eq:diverge} which diverges regardless (as long as the step size is a constant).}}
	% for the simple consensus problem.  
\end{Example}
\vspace{-0.3cm}

Second, it is challenging to deal with   {\it heterogeneous data} where the data distribution of the agents are significantly different. This is because the local update directions can be  different compared with the information communicated from the neighbors. {{Considering Example \ref{exp:diverge} again,}}
%As an example, let us again consider Example \ref{exp:diverge}. In this case, 
the divergence of DGD can be attributed to the fact that the local functions have different local data, leading to $\grd f_1(\theta) = - \grd f_2(\theta)$. 

Other practical challenges include how to implement distributed algorithms such that they scale to large networks and model sizes.  
%The above discussions highlight the challenges in dealing with non-convex problems [such as \eqref{eq:opt}] in a distributed setting. In fact, 
Moreover, an effective distributed {algorithm} should jointly design the communication and computation protocols. Addressing these challenges will be the main focus next.

%{Other challenges include, how to perform  a joint design of communication and computation protocols to make distributed algorithms as  efficient as possible? How to design algorithms that enable the agents to jointly compute high-quality, or even global optimal solutions, despite the non-convexity of \eqref{eq:opt}?} % despite the fact that the problem we    

%process {\it unbalanced} data across the agents. This scenario is especially common with \emph{federated learning} \cite{}. Concretely, we may consider the \textbf{streaming data} setting with $\pi_i(\bm{\cdot}) \neq \pi_j(\bm{\cdot})$. For this setting, the $i$th agent would never find a {stationary} solution with a small \emph{global gradient} since the latter requires data from the other agent to compute. {\color{red}[how to say this challenge more explicitly?]}

\section{Balancing Communication and Computation in Distributed Learning} \label{sec:algorithms}
\begin{figure}[t]
	\centering
	\includegraphics[width=.475\linewidth]{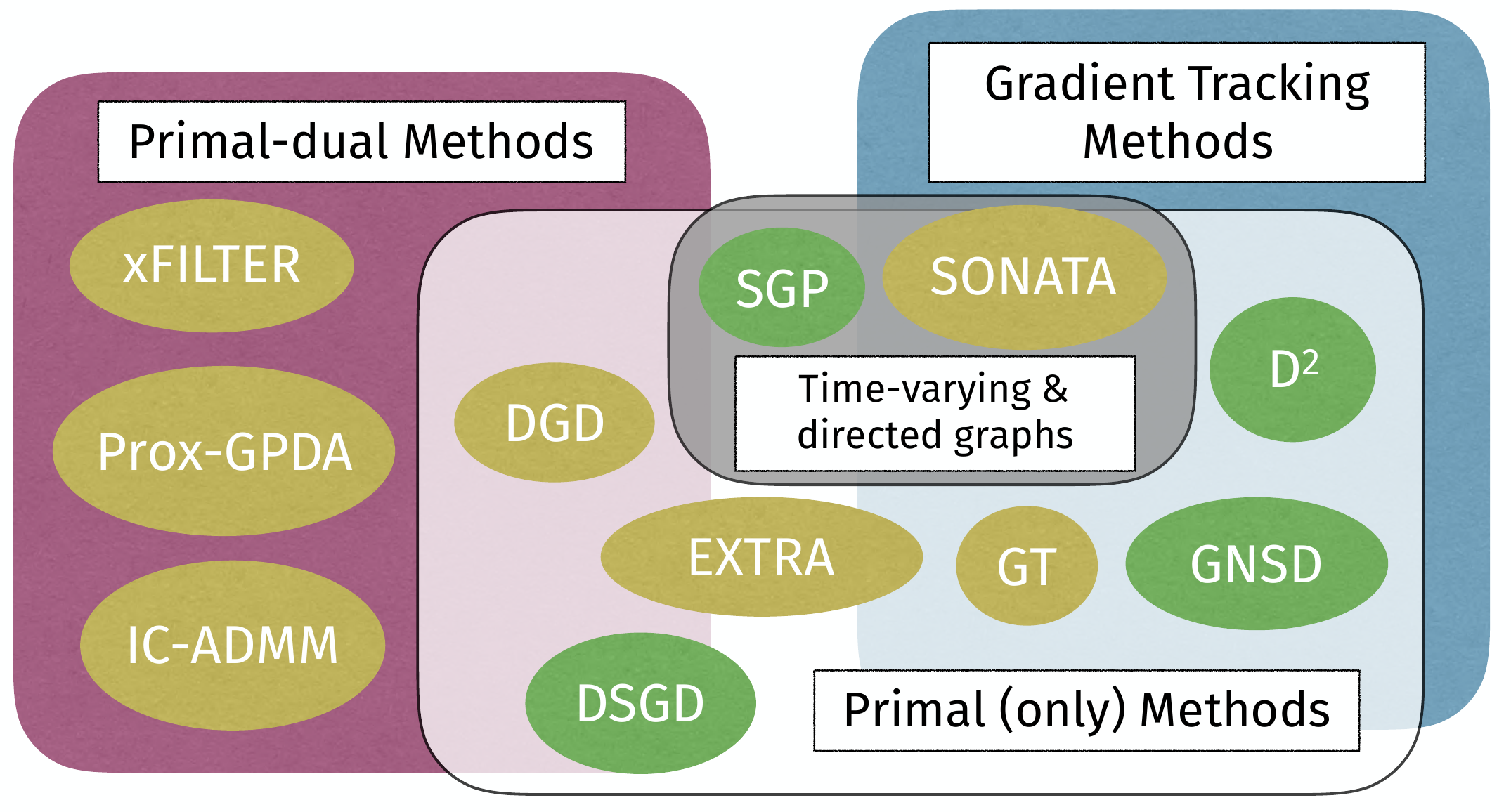}~\includegraphics[width=0.51\linewidth]{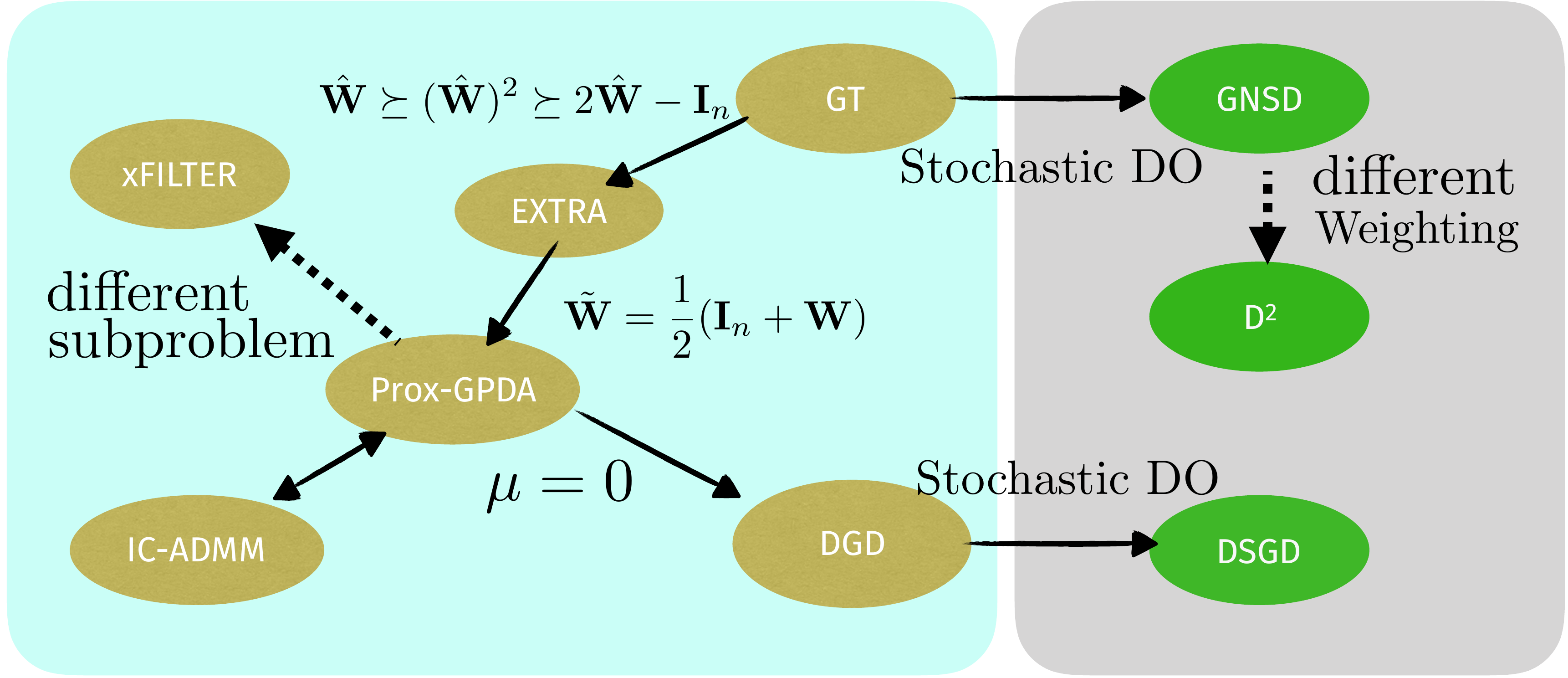}
	\vspace{-0.3cm}
	\caption{\footnotesize An overview of distributed algorithms for non-convex learning. {\color{yellow!50!black}Orange} (\resp {\color{green!50!black}green}) patches refer to algorithms designed for batch (\resp streaming) data. Line with a single arrow indicates one algorithm can reduce to another; line with double arrows means algorithms are equivalent; dotted line indicates the algorithms are related (conditions given above the line).}\label{fig:methodov}
	\vspace{-.8cm}
\end{figure}

We study distributed algorithms for tackling  problem \eqref{eq:opt}. 
{For simplicity, we assume scalar optimization variable, i.e., $d=1$, throughout this section.}
{Distributed algorithms require a balanced design for the \emph{computation} and \emph{communication} capability of a distributed learning system. This section shall delineate how existing algorithms overcome such challenge with different data oracles. In a nutshell, the \emph{batch} data setting can be tackled  using either a \emph{primal-dual optimization} framework, or a family of {\it gradient tracking} methods; while the \emph{streaming} data setting {is commonly tackled by the distributed gradient descent or  gradient tracking methods}. A summary of the reviewed algorithms can be found in Fig.~\ref{fig:methodov} (Left), and the connections between algorithms are illustrated in Fig.~\ref{fig:methodov} (Right).}
%We summarize all the reviewed algorithms in \thcedit{{\blue Fig.~\ref{fig:methodov} (Left).}} 
%\mhedit{As we will see, the \emph{batch} data setting can be tackled  using either a \emph{primal-dual optimization} framework, or a family of {\it gradient tracking} methods.} Meanwhile, the \emph{streaming} data setting \mhedit{is more commonly tackled by the latter type}. \thcedit{{\blue The connections between the algorithms will also be built in Fig.~\ref{fig:methodov} (Right).}}
%. That is, the variables are assumed to be scalars. 
Next, we review some basic concepts about distributed processing on networks.  

Considering an undirected graph $G = ({\cal V}, {\cal E})$, we define its \emph{degree matrix} as $\Db \eqdef {\rm diag}( d_1,...,d_n )$, where $d_i$ is the degree of node $i$. The \emph{graph incidence matrix} $\Ab \in \RR^{|\Ec| \times n}$ has $A_{ei}=1$, $A_{ej}=-1$ if $j>i$, $e=(i,j)\in \Ec$, and $A_{ek} = 0$ for all $k \in {\cal V} \setminus \{i,j\}$. Note that $\Ab^\top \Ab: =\Lb_G \in \RR^{n \times n}$ is the \emph{graph Laplacian} matrix. Lastly, a \emph{mixing matrix} ${\bm W}$ satisfies the following conditions:
\begin{align}\label{eq:W}
\hspace{-0.3cm}\text{\sf P1) {\rm null}}\{\Ib_n -\Wb\}=\text{{\rm span}} \{\oneb\};~\text{\sf P2)}  -\Ib_n \preceq \Wb \preceq  \Ib_n;~\text{\sf P3)}~W_{ij}=0~\text{if}~(i,j)\notin \Ec, W_{ij}>0 \; \text{o.w.}. %The mixing matrix 
\end{align}
For instance, the mixing matrix can be chosen as the doubly stochastic matrix:\vspace{-.1cm}
\beq \textstyle
W_{ij} = 1 / d_{\sf max}~\text{if}~(i,j) \in {\cal E},~~W_{ij} = 1 - d_i / d_{\sf max}~~\text{if}~i = j,~~W_{ij} = 0~\text{if}~(i,j) \notin {\cal E},\vspace{-.1cm}
\eeq
{provided that the maximum degree $d_{\sf max} := \max_{i \in {\cal V}} d_i$ is available;} see \cite{boyd2004fastest} for other designs.
For any $\param \in \RR^{n}$, we observe $\Ab^\top \Ab \param$, $\Wb \param$ can be calculated via message exchange among  neighboring agents.

{The mixing and/or graph Laplacian matrix specifies the \emph{communication} pattern in distributed learning. Since ${\bm W}^\infty = (1/n) {\bf 1}{\bf 1}^\top$, the mixing matrix allows one to compute the average distributively by repeatedly applying the mixing matrix. As the gradient $\frac{1}{n} \sum_{i=1}^n \grd f_i(\theta)$ is the average of local gradients, the easiest way to derive a `distributed' algorithm is to compute the exact average of gradient by applying ${\bm W}$ repeatedly.
	Such `distributed' algorithm will behave exactly the same as the centralized gradient algorithm, and it may save \emph{computation} (gradient or data oracle evaluation) with a faster convergence rate, yet the \emph{communication} (message exchange) cost can be overwhelming.} 

%{\color{red}We remark that an active research area is to develop distributed algorithms on \emph{stochastic} mixing matrix, suitable for \emph{directed graphs} \cite{tsianos2012push}}. 

%These matrices are important operators that enable distributed optimization.
% These matrices are important constructs which will be used widely in the distributed algorithms. 

\vspace{-0.3cm}
\subsection{Algorithms for Batch Data}\label{sub:batch}
In the batch data setting, the data oracle returns an exact gradient ${\sf DO}_i(\theta_i) = \grd f_i( \theta_i )$ at the $i$th agent at anytime. 
This setting is typical with small-to-moderate dataset where gradient computation is cheap. {To this end, the general design philosophy is to adopt techniques developed for \emph{deterministic} first-order methods and specialize them to distributed learning considering the \emph{communication} constraint.} %For instance, we will see that \emph{primal-dual based methods} provide a general framework to develop fast (and rate optimal) algorithms. 
%See the detailed account as follows.  

\textbf{Primal-Dual Methods}.
%A direct approach to distributed problems is to relate them to existing models in optimization theory. 
{Let $\param =(\theta_1,\ldots,\theta_n)^\top$ be the collection of local variables, 
	we observe that the consensus constraint $\Param$ can be rewritten as a set of linear equalities
	$\Param =  \{ \param =(\theta_1,\ldots,\theta_n)^\top ~|~\Ab \param =\zerob   \}$.
	It is then natural to consider the augmented Lagrangian (AL) of \eqref{eq:opt}}:  
\begin{align}\label{eq:AL}
\LLc(\param,\mub)=  \frac{1}{n} \sum_{i=1}^n f_i ( \theta_i ) + \mub^\top \Ab \param + \frac{c}{2}\|\Ab\param\|^2,
\end{align}
where $\mub \in \mathbb{R}^{|\Ec|}$ is the dual variable of the constraint $\Ab \param = \zerob$, and $c>0$ is a penalty parameter. The quadratic term $\|\Ab \param \|^2$ is a coupling term linking the local variables $\param = (\theta_1,...,\theta_n)^\top$.

{The proximal gradient primal-dual algorithm (Prox-GPDA) \cite{Hong17ICML} considers using a primal step which minimizes a linearized version of \eqref{eq:AL}, and the dual step which performs gradient ascent:}
%based primal-dual updates.  
\begin{align}%\label{eq: prox PDA}
\param^{t+1}   \leftarrow &\argmin_{\param \in \RR^n } \Big\{ 
\Big\langle \underbrace{\nabla \fb ( \param^t ) + \Ab^\top \mub^t + c \Ab^\top \Ab\param^t}_{=\grd_{\param} {\cal L}( \param^t, \mub^t )} , \param-\param^t \Big\rangle
+\frac{1}{2}\|\param-\param^t\|^2_{\Upsilonb+2c\Db} \Big\}, \label{eq: prox PDA S1} \\[-.2cm]
\mub^{t+1}  \leftarrow & ~ \mub^t + {c\!\!\!\!\!\underbrace{\Ab \param^{t+1}}_{=\grd_{\mub} {\cal L}( \param^{t+1}, \mub^t )},} \label{eq: prox PDA S2}
\end{align}
{where $t=0,1,...$ is the iteration number,} $\nabla\fb(\param):= \frac{1}{n} (\nabla f_1(\theta_1), \ldots,\nabla f_n(\theta_n))^\top$ stacks up the gradients, and $\Upsilonb:=\diag(\beta_1,\ldots,\beta_n)$ is a diagonal matrix.
Eq.~\eqref{eq: prox PDA S1}, \eqref{eq: prox PDA S2} lead to:
\beq \label{eq: prox PDA S1-2}
\nabla \fb ( \param^t ) + \Ab^\top \mub^t + (c \Ab^\top \Ab)\param^{t}+(\Upsilonb + 2c \Db)(\param^{t+1}-\param^t) = \zerob,~
%\eeq
%%and \eqref{eq: prox PDA S2} implies 
%\beq 
\Ab^\top \mub^{t+1}  = \Ab^\top \mub^t + c\Ab^\top \Ab \param^{t+1}, 
%\label{eq: prox PDA S2-2}
\eeq
respectively.
Setting $p_i^t \eqdef \sum_{j|(i,j)\in \Ec} \mu_j^t$ shows that \eqref{eq: prox PDA S1-2}
%, \eqref{eq: prox PDA S2-2}
%, \eqref{eq: prox PDA S2-2} 
can be decomposed into $n$ parallel updates:
\begin{equation}\label{eq: prox PDA S1-3}
\textsf{Prox-GPDA:~~}
\begin{array}{ll}
\theta_i^{t+1}  &\leftarrow 
\frac{1}{\beta_i+2c d_i}
\left\{ \beta_i \theta_i^t - \nabla f_i({\theta_i^t}) - p_i^t  
+ c \sum_{j|(i,j)\in \Ec}  (\theta_i^t + \theta_j^t) \right\},~~i=1,\ldots,n, \\
p_i^{t+1}  &\leftarrow ~p_i^t + c \sum_{j|(i,j)\in \Ec} (\theta_i^{t+1} - \theta_j^{t+1}), ~~i=1,\ldots,n. %\label{eq: prox PDA S2-3}
\end{array}
\end{equation}
{This is a distributed algorithm implementable using \emph{one} message exchange (i.e., getting $\sum_{j|(i,j)\in \Ec} \theta_j^{t+1}$) and \emph{one} DO evaluation per iteration.
	By lending to the proofs for general primal-dual algorithms, 
	\cite{Hong17ICML} shows that, under proper $c$, $\{\beta_i\}$, 
	it requires at most $\mathcal{O}(1/\epsilon)$ iterations to find an $\epsilon$-stationary solution.}

{Eq.~\eqref{eq: prox PDA S1-3} shows that a distributed algorithm with balanced \emph{computation} and \emph{communication} cost can be derived from the primal-dual method. Furthermore, the method has a strong connection with existing distributed algorithms.}
%The Prox-GPDA method is simple to implement and it has strong connection with existing distributed algorithms. 
%As we will see shortly, this leads to a fairly general algorithm that not only covers several distributed algorithms as special cases, but also provides insights into other algorithms. 
{First, we note that the inexact consensus ADMM (IC-ADMM) \cite[Algorithm 2]{chang14distributed} which applies ADMM with inexact gradient update follows exactly the same form as \eqref{eq: prox PDA S1-3}. To see the connection for other algorithms,
	let us subtract Eq.~\eqref{eq: prox PDA S1-2} at the $t$th iteration by the $(t-1)$th one}:
\beq \notag 
\grd \fb ( \param^t ) - \grd \fb ( \param^{t-1} ) + \Ab^\top (\mub^t-\mub^{t-1}) + c \Ab^\top \Ab (\param^{t} - \param^{t-1}) + (\Upsilonb+2c\Db) (\param^{t+1}- 2 \param^t + \param^{t-1}) = \zerob.
\eeq
As $\Ab^\top (\mub^t-\mub^{t-1}) = c \Ab^\top \Ab \param^{t}$, we have an equivalent form of Prox-GPDA algorithm:
\beq \label{eq: PD Gen}
\param^{t+1} = \big( \Ib_n - c ( \Upsilonb + 2c \Db )^{-1} \Ab^\top \Ab \big) ( 2 \param^t - \param^{t-1} ) - ( \Upsilonb + 2c \Db )^{-1} ( \grd \fb ( \param^t ) - \grd \fb ( \param^{t-1} ) ).
\eeq
{We see that Prox-GPDA has various equivalent forms in \eqref{eq: prox PDA S1} -- \eqref{eq: PD Gen}.}
{Below we show that a number of existing algorithms share similar communication-computation steps as Prox-GPDA.}

%\paragraph{Inexact Consensus ADMM}
%The Alternating Direction Method of Multipliers (ADMM) has been widely used for distributed optimization \cite{BoydADMMsurvey2011}. 
%To relate ADMM with the primal-dual algorithm, note that instead of \eqref{eq:consensus}, one can use the following set of linear constraints to enforce consensus:
%%ADMM is widely used for distributed optimization in the literature. 
%%To see how to connects it to the primal-dual algorithm, note that an alternative way to enforcing consensus between agents in \eqref{eq:opt} is as follows
%\begin{align}\label{eq: opt consensus matrix form ADMM}
%\ds \textsf{IC-ADMM}:~~ \min_{ \param \in \RR^n, \etab\in \mathbb{R}^{2|\Ec|} } &~ \ds \frac{1}{n} \sum_{i=1}^n f_i ( \theta_i ) ~~\text{s.t.}~~
%\theta_i = \eta_{ij},~\theta_j=\eta_{ij},~\forall (i,j)\in \Ec,
%\end{align}
%where $\etab=\{\eta_{ij}\}_{(i,j)\in \Ec} $ are the bridging variables for consensus. Since \eqref{eq: opt consensus matrix form ADMM} has two sets of variables $(\param,\etab)$, the ADMM with inexact gradient update can be applied, which optimizes  the variables in an alternating fashion. It turns out that the optimal $\etab$ is a function of $\param$, so the updates can be expressed by only using $\param$, and they take exactly the same form as Prox-GPDA in \eqref{eq: prox PDA S1-3}; see \cite[Algorithm 2]{chang14distributed}.

\emph{a) Decentralized Gradient Descent (DGD):}
Initially proposed by \cite{Nedic09subgradient} for convex problems, the DGD algorithm is one of the most popular distributed algorithms.
{This algorithm makes use of the penalized problem {$\min_{\param \in \RR^n} \sum_{i=1}^n f_i ( \theta_i ) + \frac{1}{2\alpha}\|\param\|^2_{\Ib_n-\Wb},$} where $\alpha>0$ is a penalty parameter, and applies the gradient descent method with step size $\alpha$ to yield}
%This algorithm makes use of the mixing matrix $\Wb$ satisfying \eqref{eq:W}.
%In particular, {\sf P1)} implies that if $\theta_i = \theta_j$ for any $i,j$, then $(\Ib_n - \Wb)\param=\zerob$.
%Hence it makes sense to consider a 
%penalized problem {$\min_{\param \in \RR^n} \sum_{i=1}^n f_i ( \theta_i ) + \frac{1}{2\alpha}\|\param\|^2_{\Ib_n-\Wb},$} where $\alpha>0$ is a penalty parameter.
%This insight leads one to consider tackling \eqref{eq:opt} through the penalized problem {$\min_{\param \in \RR^n} \sum_{i=1}^n f_i ( \theta_i ) + \frac{1}{2\alpha}\|\param\|^2_{\Ib_n-\Wb},$} where $\alpha>0$ is a penalty parameter. 
%Applying the gradient descent method {\blue with step size $\alpha$ to the problem} yields: %For $t=0,1.\ldots,$
\begin{align}\label{eqn: DGD}
{\sf DGD:~~}		\param^{t+1} \leftarrow  \Wb \param^{t} - \alpha \nabla \fb(\param^t), \quad \forall~t=0,1.\ldots.
\end{align}
{We note that this is a {\it primal} method, because apparently it does not involve any dual variables. Nontheless, the update formula of DGD can be derived from Prox-GPDA as we consider \eqref{eq: prox PDA S1-2} with $\mub^t=\zerob~\forall t$, $\Upsilonb+2c\Db=\alpha^{-1}\Ib_n$, and  $\Wb=\Ib_n- {c \alpha}\Ab^\top \Ab$.
	%$\Wb=\Ib_n- {\alpha}\Ab^\top \Ab$, $\Upsilonb = (1/\alpha) \Ib_n$, while setting $c=0$, $ \mub^0 = \zerob$, where the latter two conditions imply that no dual update is performed.  
	However, unlike Prox-GDPA, to guarantee convergence to a stationary solution, DGD requires a diminishing step size as $\alpha^t = 1/t$ and an \emph{additional assumption} that $\grd f_i( \theta_i )$ is bounded for any $\theta_i$ and $i=1,...,n$ \cite[Theorem 2]{Zeng19distributedGD}}.

\emph{b) EXTRA:}
The EXTRA algorithm was proposed in \cite{shi14extra} as an alternative to DGD with convergence guarantee using a constant step size.
Again, using the mixing matrix $\Wb$, the algorithm is described as follows. First, we initialize by $\param^{1} \leftarrow  \Wb \param^{0} - \alpha \nabla \fb(\param^0)$, then
%{ \begin{center}
%	\fbox{ \begin{varwidth}{\dimexpr\textwidth-20\fboxsep-3\fboxrule\relax}
\begin{align}\label{eqn: EXTRA}
{\sf EXTRA:~~}	\param^{t+1} \leftarrow  (\Ib_n+\Wb) \param^{t} - \frac{1}{2}(\Ib_n+\Wb) \param^{t-1} - \alpha [ \nabla \fb(\param^t) -\nabla \fb(\param^{t-1})],~\forall~t = 1,2,...
\end{align}
A distinctive feature of \eqref{eqn: EXTRA} is that it computes the weighted difference between the previous two iterates $\param^t$ and $\param^{t-1}$.
Interestingly, the above form of EXTRA is a special instance of Prox-GPDA. Setting $\Upsilonb + 2c \Db = \alpha^{-1} \Ib_n$ in \eqref{eq: PD Gen}, we obtain 
\begin{align}\label{eqn: EXTRA equ}
\param^{t+1} \leftarrow  (2 \Ib_n - 2 c\alpha \Ab^\top \Ab) \param^{t} 
- \frac{1}{2} (2 \Ib_n - 2 c\alpha \Ab^\top \Ab)\param^{t-1}   - \alpha [ \nabla \fb(\param^t) -\nabla \fb(\param^{t-1})]. 
\end{align}
Choosing $\Wb=\Ib_n - 2 c \alpha \Ab^T\Ab$ and the above update recovers \eqref{eqn: EXTRA}. The original proof in \cite{shi14extra} assumes convexity for \eqref{eq:opt}, 
but due to the above stated equivalence, the proof for Prox--GPDA for non-convex problems carries over to the EXTRA algorithm. %\mhcomment{we can also highlight that this connection is unknown, in the response}
{It is worth mentioning that the original EXTRA algorithm in \cite{shi14extra} takes a slightly more general form. That is, the term $\frac{1}{2}(\Ib_n+\Wb)$ in \eqref{eqn: EXTRA} can be replaced by another mixing matrix $\tilde{\Wb}$, which satisfies
	$\frac{\Ib_n+{\Wb}}{2}\succeq \tilde{\Wb}\succeq {\Wb}$, and ${\rm null} \{ \Ib_n-\tilde{\Wb}\}= {\rm span}\{ {\bf 1} \}$. However, it is not clear if this general form works for the non-convex distributed problems.}

\emph{c) Rate Optimal Schemes:}
A fundamental question about distributed problem \eqref{eq:opt}  is: ``\emph{what are the minimum computation and communication cost required to find an $\epsilon$-stationary solution?"}
{An answer to this question is given in \cite{sun18optimal}. For any distributed algorithm using gradient information, it requires at least $ \Omega \big( (\epsilon \sqrt{ \xi(\Lb_G)} )^{-1} \big)$ communication rounds, {and $\Omega \big( \epsilon^{-1} \big)$} rounds of gradient evaluation\footnote{{In each gradient evaluation, each local node $i$ evaluates $\nabla f_i(\cdot)$ once.}} to attain an $\epsilon$-stationary solution, where $\xi(\Lb_G) \eqdef \frac{ \lambda_{\min}(\Lb_G)}{\lambda_{\max}(\Lb_G)}$ is the ratio between the smallest non-zero and the largest eigenvalues of the graph Laplacian matrix $\Lb_G$.} %$\Lb_G=\frac{c}{L}\Ab^\top \Ab $ and \mhedit{$\xi(\Lb_G)$ is }.}
Interestingly, Prox-GPDA, EXTRA and IC-ADMM achieve the lower communication and computation bounds in star or fully connected networks.
For general network topology,  \cite{sun18optimal} proposed a {\it near optimal scheme} called xFILTER, which updates $\param$  by considering: %Specifically, it considers the following subproblem, %The xFILTER algorithm updates  $\param$ by using the following update: 
\begin{align}%\label{eq: prox PDA}
\param^{t+1} = \argmin_{\param \in \RR^n } \left\{ \nabla \fb ( \param^t )^\top (\param-\param^t) + \mub^\top \Ab\param + \frac{c}{2}\|\Ab\param\|^2
+\frac{1}{2}\|\param-\param^t\|^2_{\Upsilonb} \right\}. \label{eq: xFILTER S1} 
%\\
%\mub^{t+1}  \leftarrow & \mub^t + \Sigmab\Ab \param^{t+1}, \label{eq: xFILTER S2}
\end{align}
{Compared to \eqref{eq: prox PDA S1}, the quadratic term $ \frac{c}{2}\|\Ab\param\|^2$  is not linearized}. This term couples the local variables, so \eqref{eq: xFILTER S1} itself does not lead to a distributed update for $\param^{t+1}$. To resolve this issue, the authors proposed to generate $\param^{t+1}$ by using the $Q$th order Chebychev polynomial to approximately solve \eqref{eq: xFILTER S1}.
They showed that setting $Q={\widetilde{\cal{O}}} \big( 1/ \sqrt{ \xi(\Lb_G)} \big)$ suffices to produce an algorithm that requires  $\widetilde{\mathcal{O}}\big( (\epsilon \sqrt{ \xi(\Lb_G)} )^{-1} \big)$ communication rounds { and ${\mathcal{O}}(\epsilon^{-1})$ gradient evaluations rounds}, where the notation ${\widetilde{\mathcal{O}}}(\cdot)$ hides a $\log$ function of $n$ (which is usually small). {This matches the aforementioned lower bounds.} Similarly, the communication effort required is near-optimal (up to a multiplicative logarithmic factor). {Further, by comparing with all other batch methods (to be) reviewed in this work, this is the only algorithm whose gradient evaluation complexity is {\it independent} of the graph structure.}

\textbf{Gradient-Tracking Based Methods}.
Another class of algorithms that can deal with the non-convex problem \eqref{eq:opt} leverages the technique of {\it gradient tracking} (GT). The method is based on the simple idea that, if every agent has access to the global gradient ${1}/{n}\sum_{i=1}^n \nabla  f_i(1/n\sum_{j=1}^n \theta^t_j)$, then the (centralized) GD can be performed at each agent. 
%But this requires that the local agents  have access to the averages of the local gradients and the local parameters. 
The GT technique provides {an iterative approach} to do so approximately. The algorithm performs {two message exchanges each iteration} [with $\hWb$ satisfying \eqref{eq:W}]:
\begin{equation}
\begin{array}{ll}
{\sf GT}:~&\param^{t+1} \leftarrow  \hWb \param^{t} - \alpha \gb^t,  \label{eq: GT S1}~~ \gb^{t+1} \leftarrow \hWb \gb^{t} + \grd \fb(\param^{t+1}) - \grd \fb(\param^{t}),\; \forall~ t=1,2, \ldots.         \end{array}
\end{equation}
The $i$th element $g^t_i$ of $\gb^t$ is the local estimate of the global gradient at each agent $i$, obtained by mixing the estimates of its neighbors and refreshing its local $\nabla f_i$. As shown in \cite{scutari2019distributed}, the GT algorithm \eqref{eq: GT S1} converges at a rate of ${\cal O}(1/\epsilon)$ to a stationary point. One key strength of the GT based methods is that they can also work in directed and time-varying graphs; see \cite{scutari2019distributed} for more discussions.

{We remark that the GT based method is related to the general form of EXTRA. %and the primal-dual type methods do not include each other as special cases.  
	To see this, we subtract the updates of $\param^{t+1}$ and $\param^{t}$,     and apply the update of $\gb^{t}$ to obtain
	\begin{align} \label{eq: GT Rewrite}
	\param^{t+1} \leftarrow 2\hWb\param^{t}-\hWb^2\param^{t-1}-\alpha \left(\nabla \fb (\param^{t})-\nabla \fb (\param^{t-1})\right),~\forall~t=1,2,....
	\end{align}
	It can be shown that if  $\hWb$ satisfies $\hWb\succeq (\hWb)^2\succeq 2 \hWb -\Ib_n$, then the algorithm takes the same form as the generalized EXTRA discussed after \eqref{eqn: EXTRA equ}; see \cite[Section 2.2.1]{nedic2017achieving}. However, the analysis for generalized EXTRA only works on convex problems, and does not carry to the GT method in the non-convex setting.}

{We remark that all the above algorithms converge for the challenging \Cref{exp:diverge} except for the DGD algorithm. This is because for the latter example, the gradient can be unbounded.}
%{\blue Tsung-Hui: should we mention this after (22), right after the DGD since Example 3 illustrates the failure of a DGD like algorithm?}

%equation in \eqref{eqn: EXTRA equ}, yet the GT method is \emph{not} a special case of Prox-GPDA since i) there is an additional $\Wb$ term in front of the gradients, ii) it is not possible to find a matrix $\Wb'$ satisfying \eqref{eq:W}, $2\Wb = (\Ib_n+\Wb')$ and  $\Wb^2 = \frac{1}{2}(\Ib_n+ \Wb')$ simultaneously.

\vspace{-0.3cm}
\subsection{Algorithms for Streaming Data} \label{sub:streaming}
{In the streaming data setting, the data oracle returns ${\sf DO}_i ( \prm^{t}_i)$ which is an unbiased estimator of $\nabla f_i(\prm^{t}_i)$ with finite variance under Assumption~\ref{ass:variance}. %For simplicity, we assume $d=1$ throughout this section. 
	This data model is typical in processing large-to-infinite datasets. 
	%It is relevant when the data is streaming in, or when computing the full gradient is expensive.
	In this setting, balancing between \emph{communication} and \emph{computation} cost is an important issue since even the centralized algorithm may have slow convergence. The first study of distributed stochastic algorithm dates back to Tsitsiklis et al. \cite{tsitsiklis86} which studied the asymptotic convergence of the DSGD algorithm reviewed below. 
	The DSGD algorithm is relevant to the distributed estimation problem important in adaptive signal processing, as such, many works are devoted to studying its transient behavior (of bias, mean-squared error, etc.), e.g., \cite{cattivelli2009diffusion,kar2012distributed} and the overview in \cite{sayed2013diffusion}. Unfortunately, these works are mainly focused on \emph{convex problems}. Below, we review the more recent results dedicated to the non-convex learning setting with non-asymptotic convergence analysis.} 
%\mhdelete{Owing to that, \emph{stochastic} oracles are often used to model the data acquisition process. Centralized stochastic algorithms have enjoyed great practical successes despite the need to deal with randomness in the data.} 
%\mhedit{ The history of design and analysis of distributed stochastic algorithms dates back to the seminal work of Tsitsiklis, Bertsekas and Athans \cite{tsitsiklis86}. 
%However, extending the modern convergence rate analysis of batch distributed algorithms to this setting is not straightforward. This is because with a stochastic data oracle, the step size used in algorithms needs to be carefully {controlled},  and one has to pay special attention when the {\it distributions} of data are different across  agents.}

\emph{a) Distributed Stochastic Gradient Descent (DSGD):}
{This class of algorithm  replaces the deterministic oracle in the DGD with the stochastic oracle \eqref{eq:stream}. It takes the following form:
	%From a global view, the iterates of D-PSGD can is shown in the following
	\begin{equation}
	{\sf {DSGD}:~~}\param^{t+1} \leftarrow \Wb\param^{t}-\alpha^{t} {\sf DO} ( \param^{t}),\label{eq.gp} %\hnabla_{\param} {\Fb}(\param^{(t)},\mathbf{\xi}^{(t)}),\label{eq.gp}
	\end{equation}
	where $\alpha^t > 0$ is the step size and we defined ${\sf DO} ( \param^{t}) \eqdef ({\sf DO}_1 ( \theta_1^{t}), ..., {\sf DO}_n (\theta_n^t) )^\top$. Obviously, DSGD can be implemented in a distributed manner via the mixing matrix $\Wb$.  
	The study of such an algorithm in the non-convex setting dates back to the work \cite{tsitsiklis86}. Among other results,  the authors showed that if the step size sequence satisfies $\alpha^t \leq c/t$ for some $c >0$, the DSGD algorithm converges almost surely to a first-order stationary solution.
	However, \cite{tsitsiklis86} mainly provides asymptotic convergence conditions, without a clear indication of whether DSGD can outperform its centralized counterpart.}
%Subsequently, there has been a vast literature on analyzing and applying DSGD or its variants. 
%For example, the work \cite{bian13} studied a constrained version of DSGD.
%; the work
%	\cite{cattivelli2009diffusion} developed variants of DSGD and applied them to an important class of distributed estimation problems arising in adaptive signal processing.
%	For an excellent review for the line of works about distributed estimation, we refer the readers to the overview paper \cite{sayed2014adaptive}.} %The algorithm replaces the full gradients used in DGD by their stochastic counterpart:

%It is shown in \cite{bian13} that under some qualification assumptions, DSGD converges to a stationary point of the {\blue non-convex finite sum} problem as $t \rightarrow \infty$.
Recently, DSGD [a.k.a.~decentralized parallel stochastic gradient descent (D-PSGD)] has been applied for decentralized training of neural networks in \cite{Jiang2017}, and the convergence rate has been analyzed in \cite{Lian17}.
In the analysis by \cite{Lian17}, the following condition on the data across agents is assumed: %\mhcomment{say this condition does not hold for quadratic}:
\begin{equation}\label{eq:assda}
\textstyle n^{-1} \sum_{i=1}^n | \grd f_i ( \theta )- \grd f(\theta )|^2 \le \varsigma^2 < \infty,~\forall~\theta \in \RR.
\end{equation}
{Such an assumption can be difficult to verify, and it is {\it only} required when analyzing the convergence rate of DSGD for non-convex problems}. For example, if the loss function is quadratic, then the corresponding gradient is a linear function of $\theta$, e.g., $\grd f_i( \theta ) = a_i \theta + b_i$, $i=1,...,n$. The LHS of \eqref{eq:assda} is unbounded if $a_i \neq (1/n)\sum_{j=1}^n a_j$, \ie whenever the cost function is heterogeneous. 

%It is interesting to see that Example \ref{exp:diverge} in Sec. \ref{sub:challenge} is an extreme case where the data are heterogenous across the agents. Clearly condition \eqref{eq:assda} is violated here, and DSGD does not behave well in this case.

Under \eqref{eq:assda} and Assumption \ref{ass:gen} \& \ref{ass:variance}, for any {\it sufficiently large} $T$, if we set $\alpha^t = {\cal O}( \sqrt{n/ (\sigma^2 T)})$ for all $t \geq 0$, then the DSGD finds an approximate stationary solution to \eqref{eq:opt} satisfying $\EE[ {\sf Gap}( \param^{\tilde{t}} ) ] = {\cal O}( \sigma / \sqrt{nT} )$, where $\tilde{t}$ is uniformly drawn from $\{1,...,T\}$ \cite[Corollary 2]{Lian17}.  
{Compared to the centralized SGD algorithm, a speedup factor of $1/\sqrt{n}$ is observed}, which is due to the variance reduction effect by averaging from $n$ nodes. Yet achieving  this requires $\varsigma^2 = {\cal O}(1)$ such that the data is {\it homogeneous} across the agents. {Also, see \cite{vlaski2019distributed,vlaski2019distributedb} which show that the DSGD algorithm converges to a second-order stationary solution under a similar condition as \eqref{eq:assda}.}

{In summary, the DSGD algorithm is simple to implement, but it has a major limitation when dealing with heterogeneous data. Such a limitation will also be demonstrated in our numerical experiments.}

%\mhcomment{Also another point of this paper is that decentralized algorithm can outperform the centralized algorithm when $N$ is large. We should have comments on this. }.

%{\red[do other algorithms I mentioned above require this condition?]} {\red[can you say why this is needed intuitively, and is it needed due to insufficient in analysis, or fundamentally it is required? If this assumption is not satisfied, can we build an easy (numerical) example to show that D-PSGD fails?]}

%Note that choosing a shrinkage step-size is necessary, since in stochastic settings of the algorithms, the estimate of the gradients is not accurate at each step, so ensuring the convergence of the iterates to stationary points requires the decrease of the step-size which is similar as the centralized case of performing SGD.

\emph{b) D$^2$ Algorithm:} {To relax the local data assumption \eqref{eq:assda} from DSGD, an algorithm named D$^2$ has been proposed in \cite{Tang18}.} %The main difference is that D$^2$ can get ride of assumption such that it has a better performance in the case where the nodes have the heterogeneous data.
Again, using the mixing matrix $\Wb$, the recursion of D$^2$ is given as:
\begin{equation}\label{eq:d2}
{\sf D^2:~~} \param^{t+1} \leftarrow 2 \Wb \param^{t}- \Wb \param^{t-1}-\alpha^{t} \Wb \left(\sf{DO}(\param^{t})-\sf{DO}(\param^{t-1})\right),~\forall~t \geq 0.%\notag
\end{equation}
In addition to the previous conditions on the weight matrix \eqref{eq:W}, D$^2$ requires a special condition $\lambda_{\min}(\Wb)>-1/3$. Basically, the condition implies that the weight of combining the current node is greater than the ones of combining its neighbors. Together with Assumption \ref{ass:gen} \& \ref{ass:variance}, for any sufficiently large $T$, we set $\alpha^t = {\cal O}( \sqrt{n/ (\sigma^2 T)})$ for all $t \geq 0$, and D$^2$ finds an approximate stationary solution \cite{Tang18} to \eqref{eq:opt} satisfying $\EE[ \| n^{-1} \sum_{j=1}^n \grd f_j( \overline{\prm}^{\tilde{t}})\|^2 ] = {\cal O}( \sigma / \sqrt{nT} )$, where $\tilde{t}$ is uniformly drawn from $\{1,...,T\}$. %\mhdelete{However, we note that \cite{Tang18} did not provide the convergence rate of the consensus error.} %\mhcomment{add discussion about comparison with centralized algorithms.}

%the D$^2$ finds an $\epsilon$-stationary point in $\mathcal{O}(\sigma^2/ n \epsilon^2)$ iterations (when $\epsilon$ is sufficiently small) \cite{Tang18}.  

%\thcdelete{It is shown in \cite{Tang18} that when Assumption \ref{ass:gen}, \ref{ass:variance} and the condition $\lambda_{\min}(\Wb)>-1/3$ are satisfied, then D$^2$ converges to an $\epsilon$-stationary (see footnote$^1$) solution of problem \eqref{eq:opt} in $\mathcal{O}(\sigma^2/ n \epsilon^2)$ iterations. }

{In fact, comparing \eqref{eq:d2} to \eqref{eq: GT Rewrite} reveals a close similarity between D$^2$ and  GT: both algorithms use the current and the previous ${\sf DO}$'s, and both require {\it two} local communication rounds per iteration.} The difference is that the GT method applies a squared mixing matrix $\Wb^2$ on $\param^{t-1}$ instead of the mixing matrix $\Wb$ for D$^2$, and there is a $\Wb$ multiplying the difference of the gradient estimates. %\mhdelete{It is therefore natural to consider analyzing an algorithm based on the actual GT method.} 
{Such a seemingly minor difference turns out to be one major limiting factor for D$^2$; see the example below}. %In particular, it fails to converge for very simple choices of weight matrix $\Wb$.}
{
	\vspace{-0.3cm}
	\begin{Example}\cite{zhang2019decentralized} \label{exp:diverge:2} Consider a line network consisting of three nodes, with $f_i(x) = (x-b_i)^2, \; i=1,2,3$ (for some fixed $b_i$), and mixing matrix: $\Wb =[0.5, 0.5, 0; 0.5, 0, 0.5; 0, 0.5, 0.5]$, which has eigenvalues $\{-0.5, 0.5, 1\}$. %:{\small
		%	\begin{align}
		%	\Wb = \left( \begin{array}{ccc}
		%	0.5 & 0.5 & 0\\
		%	0.5 & 0 & 0.5\\
		%	0 & 0.5 & 0.5
		%	\end{array} \right).
		%	\end{align}}
		One can show D$^2$ diverges for any constant $\alpha^t\le 0.25$, or diminishing step size $\alpha^t=1/t$. 
	\end{Example}
	\vspace{-0.3cm}}

\emph{c) Distributed Stochastic Gradient Tracking:}
%\noindent{Gradient-Tracking based Non-convex Stochastic Algorithm for Decentralized nonconvex problems (GNSD)}: 
{How to design algorithms that can deal with heterogeneous data, while requiring conditions weaker than that of D$^2$?} An algorithm called Gradient-tracking based Non-convex Stochastic algorithm for Decentralized training (GNSD) has been proposed in \cite{luho19}, which is essentially a stochastic version of the GT method in \eqref{eq: GT Rewrite}:
%\begin{equation}
%\begin{array}{ll}
%{\sf GNSD}:~&\param^{t+1} \leftarrow  \Wb \param^{t} - \alpha^t \gb^t,  \label{eq: GNSD 1}~~ \gb^{t+1} \leftarrow \Wb \gb^{t} + \sf{DO}(\param^{t+1})-\sf{DO}(\param^{t}),
%\end{array}
%\end{equation}
%The above update is equivalent to the following recursion:
\begin{equation}\label{eq:gnsd}
{\sf GNSD}:~\param^{t+1} \leftarrow 2\Wb\param^{t}-\Wb^2\param^{t-1}-\alpha^{t}\left(\sf{DO}(\param^{t})-\sf{DO}(\param^{t-1})\right).
\end{equation}
It is shown that GNSD has the similar convergence guarantees as D$^2$, without requiring the assumption \eqref{eq:assda} and the condition $\lambda_{\min}(\Wb)>-1/3$. %\mhedit{However, it is worth noting that both GNSD and D$^2$ requires {\it two} communication rounds per iteration $t$ (), while the DSGD only requires one. }
%\todelete{In particular, when $\Wb$ in \eqref{eq:gnsd} is chosen as $\Ib$, then \eqref{eq:gnsd} is exactly the same as the update of %$\param^{(t+\frac{1}{2})}$ shown in \eqref{eq:d2}. Note that weighting matrix in \eqref{eq.yup} is critical in ensuring convergence of GNSD, which gives the contraction property of $\|\param^{(t)}-\1\bar{\param}^{(t)}\|$ since $\param^{(t)}-\1\bar{\param}^{(t)}\in\textrm{col}(\Wb)$, where $\bar{\param}^{(t)}:=\1^T\param^{(t)}/n$ and $\1$ denotes an all ones vector.} %{\red[I don't understand this; for any algorithm the weight matrix is important.]} {\red[intuitively why gradient tracking can also deal with heterogeneous data?]}
%Under \eqref{eq:W}, Assumptions \ref{ass:gen} and \ref{ass:variance}, for any sufficiently large $T$, we set $\alpha^t = {\cal O}( \sqrt{n/ (\sigma^2 T)})$ for all $t \geq 0$, and GNSD finds an approximate stationary solution to \eqref{eq:opt} satisfying $\EE[ {\sf Gap}( \param^{\tilde{t}} ) ] = {\cal O}( \sigma / \sqrt{nT} )$, where $\tilde{t}$ is uniformly drawn from $\{1,...,T\}$.
%and setting the step size as $\alpha^t = \mathcal{O}(1/\sqrt{n\times t})$, the GNSD iterates converge to an $\epsilon$-stationary point of \eqref{eq:opt} in $\mathcal{O}(\sigma^2/n \epsilon^2)$ iterations (when $\epsilon$ is sufficiently small).

{To summarize, D$^2$ and GNSD address the challenge of heterogeneous data unique to the streaming data setting, while simple methods such as DSGD require data to be homogeneous. On the other hand, D$^2$ and GNSD require additional communication per iteration compared with DSGD. We remark that there appears to be no work extending primal-dual type algorithm/analysis to the streaming setting.}\vspace{-.3cm}

\subsection{Other Distributed Algorithms}\vspace{-.3cm}
Despite the differences in DOs used and assumptions needed for convergence, the reviewed algorithms may be regarded as variants of unconstrained gradient descent methods for a single parameter (vector) on a fixed communication graph. However, special communication and computation architectures may arise in practice. Here we conclude the section by highlighting a few works in relevant directions.
% consider directed and/or time-varying graph topologies, which commonly arise in practice. 
% for example, when the communication channels are unreliable and transmit data only in one direction. 

\paragraph{Coordinate Descent Methods} When the optimization model \eqref{eq:opt} involves multiple variables, it is often beneficial to adopt a \emph{coordinate descent} method, which optimizes only one variable at a time, holding the others as constant. An example is \emph{matrix factorization} problem discussed in Example~\ref{exp:mat}. In specific, \cite{Hong17ICML,daneshmand2018decentralized} respectively proposes to combine Prox-GPDA, GT with coordinate descent to tackle the distributed dictionary learning problem (\emph{batch} data), with convergence guarantee. %Other related works include \

%	\paragraph{Constrained/Non-Smooth Problems} The model \eqref{eq:opt} may include constraints or (non-smooth) regularization which induces desirable properties such as sparsity on the solutions. This can be described in our model by writing $f_i(\prm) = h_i(\prm) + g_i(\prm)$, where $h_i, g_i$ are respectively smooth and non-smooth functions. Distributed algorithms developed along this direction include \cite{Zeng19distributedGD,scutari2019distributed,wai2017decentralized}.
%	In particular,
%	\cite{Zeng19distributedGD} studies a proximal version of DGD,  \cite{scutari2019distributed} applies GT and successive convex approximation, \cite{wai2017decentralized}  combines GT with projection-free (a.k.a.~Frank-Wolfe) optimization.
%	However, these works only deal with  situations where the agents have a {\it common} constraint and/or a non-smooth term in the local cost function. Developing distributed algorithms with heterogeneous non-smooth terms and constraints is an interesting direction.
\paragraph{Directed and Time Varying Graphs}
{Throughout this paper, we have assumed that the graph connecting the agents is undirected and static. However, directed and/or time-varying graph topology may arise  in practice, e.g., with unreliable network. Several works have been proposed for various settings.}
For \emph{batch data}, \cite{scutari2019distributed} proposed the SONATA algorithm which combines GT with PushSum technique; for \emph{streaming data}, \cite{assran19} proposed the Stochastic Gradient Push (SGP) algorithm which combines SGD and PushSum technique. Both SONATA and SGP are shown to {converge sublinearly to a stationary solution on time-varying and directed graphs.} %, and analysis have demonstrated that they both converge sublinearly to a stationary solution. 

%	{\blue Tsung-Hui: where to mention online algorithms? }

%{\color{red}We remark that an active research area is to develop distributed algorithms on \emph{stochastic} mixing matrix, suitable for \emph{directed graphs} \cite{tsianos2012push}}. 

%\thccomment{Will we discuss algorithms that can handle problem (1) with constraints or with composite objective function? For example, prox-DGD can handle convex constraints by Zhang\&Yin's paper; Prox-GPDA/IC-ADMM can be extended to that with additional linear constraint by Zhang Jiawei's paper; some GT methods by Aldo can handle problem with constraints as well}

%	\begin{table}
%		\caption{Summary of state-of-the-art algorithms}
%		
%		{\small
%			\begin{tabular}{|c |c | c | c |c|}
%				\hline
%				\backslashbox{$\mbox{Algorithms}$}{$\mbox{Oracles}$} & batch data & streaming data & comments
%				xxx data \\
%				\hline
%				Primal-Dual &  & &  &\\
%				\hline
%				ADMM &  & & &\\
%				\hline
%				Gradient Tracking &  & & &\\
%				\hline				
%				ADMM &  & & &\\
%				\hline
%				\hline
%			\end{tabular} } \label{tableCompare}
%		\end{table}

\vspace{-.3cm}
\section{Practical Issues and Numerical Results} \label{sec:practical}
\vspace{-.3cm}

We discuss the practical issues related to the implementation of distributed algorithms. We aim to demonstrate how system and algorithm parameters, such as network size, computation/communication speed, batch size and model size, should be considered jointly to decide on the most suitable  algorithm. In particular, we compare their effects on the overall runtime performance of algorithms.

\begin{figure}[t]
	\centering
	\hspace{-0.3cm} \includegraphics[width=.37\linewidth]{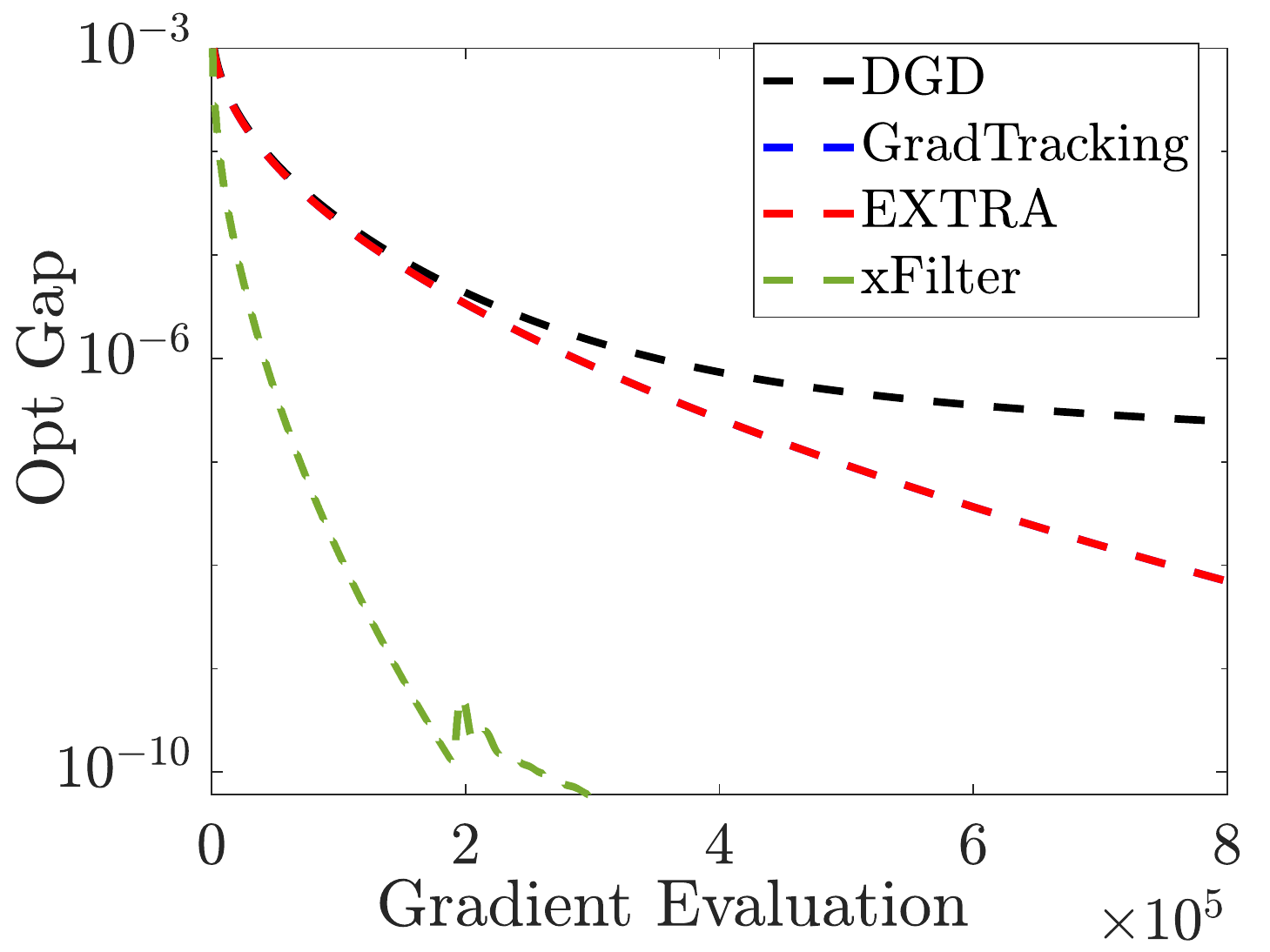}\quad~
	\resizebox{.465\linewidth}{!}{\small
		\begin{tabular}[b]{l llll lll}
			\toprule
			{\bf Cost per iter.} & \multicolumn{4}{l}{\bf Computation ($m$)} & \multicolumn{3}{l}{\bf Communication  ($n$)} \\ \midrule
			Settings & 128 & 8 & 64 & 256 & 2 & 8 & 32  \\ \midrule
			MSI, DSGD & 1 & & & & 0.30 & 2.38 & 8.87 \\
			MSI, GNSD & 1 & & & & \cellcolor{red!10}0.64 & \cellcolor{red!10}4.78 & \cellcolor{red!10}19.4 \\
			AWS, DSGD & 1 & & & & 0.14 & 1.47 & 4.12 \\
			AWS, GNSD & 1 & & & & \cellcolor{green!10}0.17 & \cellcolor{green!10}1.60 & \cellcolor{green!10}4.21 \\ \midrule
			MSI, DSGD & & 1& \cellcolor{blue!10} 1.09 &  \cellcolor{blue!10} 1.36 & &  & 2.61 \\
			MSI, GNSD & & 1& ~1.12 & ~1.45 & &  & 8.47 \\\bottomrule
	\end{tabular}}
	\caption{\footnotesize (Left) Stationarity gap against iteration number of different algorithms with a synthetic dataset and $n=32$ agents. Note that the curve for Gradient Tracking and EXTRA overlaps with each other. (Right) Normalized running time \emph{per iteration / message exchange round} on the MSI and AWS clusters under different settings for batch size $m$ and network size $n$.}\label{fig:matlab_alg}\vspace{-.5cm}
\end{figure}

Our experiments are conducted on two different computer clusters, one provided by Minnesota Supercomputing Institute (MSI), another by Amazon Web Services (AWS). 
%\todelete{The major differences between these two systems are that, the MSI machines only have CPUs and have worse inter-node communication capabilities, while the AWS nodes have both CPUs and GPUs, and they have high-speed inter-node communication capabilities}
The MSI cluster has better independent computation power at each node, but worse communication bandwidth than the AWS cluster; see Fig.~\ref{fig:matlab_alg} (right). Specifically, MSI nodes have Intel Haswell E5-2680v3 CPUs at 3.2 GHz, 14Gbps inter-node communication, while AWS nodes have Intel Xeon E5-2686v4 CPUs at 3.0 GHz, NVIDIA K80 GPUs and 25Gbps inter-node communication. 
% Running the algorithms on different systems showcase how system capabilities can affect performance of distributed algorithms on machine learning tasks.
%{that MSI has a worse communication compared with its computation speed and AWS has a worse computation compared with its communication speed.}}

Two sets of experiments are conducted. The first set compares different algorithms on a single machine. Since the distributed implementation is only simulated, the purpose of this set is to understand the theoretical behavior of algorithms. The second set of experiments showcases the algorithm performance on truly distributed systems. These algorithms are implemented in Python 3.6 with the MPI communication protocol. 
We benchmark the algorithms by using the gap ${\sf Gap}(\param)$ in \eqref{eq:stationarity}. 
%{\red [also mention either here or under each figure the stopping criteria].}%our numerical results in the following will}%We emphasize again that At this point, let us note that . It may not be fair to compare the performance of DSGD and GNSD since xxx %and the 

\paragraph{Experiment Set I}  %by counter the required number of iterations. 
We consider tackling a regularized logistic regression problem with a \emph{non-convex} regularizer in a distributed manner. We use similar notations as in Example 1, \ie the feature is $\xb_i^{\ell}$ and the label is $y_i^{\ell}$. Let $\lambda, \rho > 0$ be the regularizer's parameters, each local cost function $f_i$  is given by
\begin{align*} \textstyle
f_i(\thetab_i) = \frac{1}{M_i}   \sum_{\ell=1}^{M_i} \log{\left( 1+\exp(-y_{i}^{\ell} \thetab_i^\top {\xb}_{i}^{\ell})\right) } + \lambda \sum_{s=1}^d \frac{ \rho \theta^2_{i,s}}{1+\rho \theta_{i,s}^2}.
\end{align*}  
%\mhdelete{Note that the non-convex regularizer promotes sparsity in the solution.} 
All algorithms are implemented in MATLAB. We set the dimension at $d=10$, and generate $M_i = 400$ synthetic data points on each of the  $n=32$ agents; the communication network is a random regular graph of degree $5$. 
{The stationarity gap versus iteration number for the surveyed {batch algorithms} is shown in Fig.~\ref{fig:matlab_alg} (left). As seen, in terms of total number of full gradient evaluations, %the batch algorithms consistently outperform streaming algorithms (batch size $m_t=40$), a
	the xFILTER is the fastest. The observation we made is also consistent with the theoretical prediction, because the discussion in Sec. \ref{sub:batch} suggests that xFILTER is the only algorithm whose total gradient evaluation is {\it independent} of the graph structure, and matches the centralized GD.}%We conclude that for small dataset where a batch data oracle is available, the batch algorithms based on primal-dual methods should be employed for the best performance.

\paragraph{Experiment Set II}
We focus on  the DSGD  and GNSD algorithms for streaming data, and apply them to train a neural network as in Example 1, and the task to classify handwritten digits from the MNIST dataset. The neural network contains two hidden layers with 512 and 128 neurons each, and $4.68\times10^5$ parameters in total. The training data set has $4.8\times 10^4$ entries and is divided evenly among $n$ nodes. The DSGD and GNSD algorithms adopt the streaming data oracle in Sec.~\ref{sec:framework}), and all agents use the same mini-batch sizes $m_t=m$. The communication graph is a random regular graph with degree $5$.

Before we compare the overall performance of different algorithms, we first examine the computation/communication performance for our two clusters in running DSGD/GSND. In the upper part of Fig.~\ref{fig:matlab_alg} (right), we compare the \emph{relative} computation and communication costs on MSI and AWS. It is clear that the AWS cluster has better communication efficiency compared to the MSI. For example, consider running GNSD on a network with $n=8$ nodes, and  set the computational time per iteration as 1 unit of time. Observe that AWS uses 1.6 units of time on communication, while MSI uses 4.78 units. 

%For each scenario, the algorithms are stopped whenever the data set is passed by a fixed number of times. %{\red [epoch]}%Unless otherwise stated, the 
%{\red[balanced, unbalanced?]}

\begin{figure}[t]
	\centering
	\includegraphics[width=.28\linewidth]{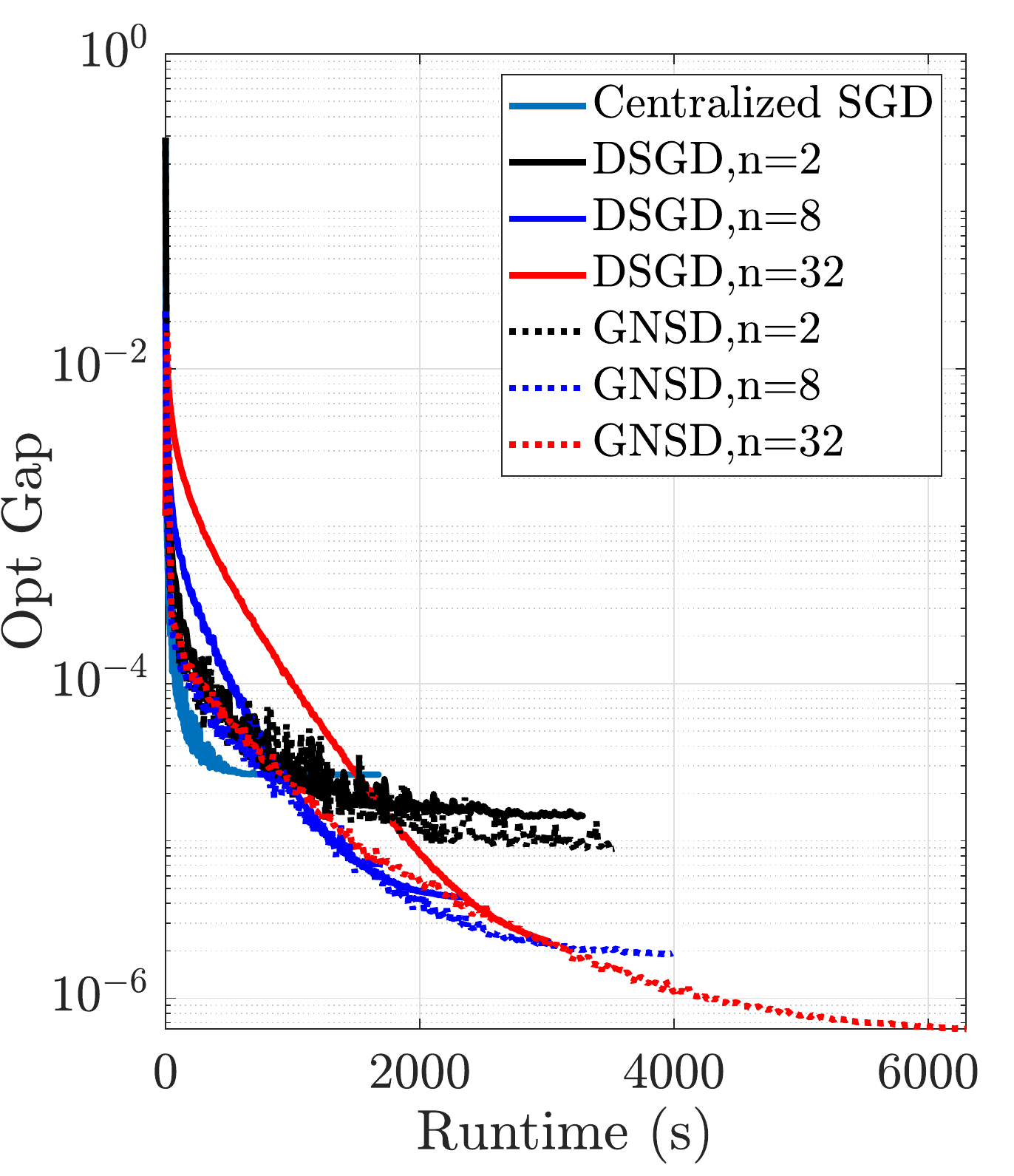}~
	\includegraphics[width=.28\linewidth]{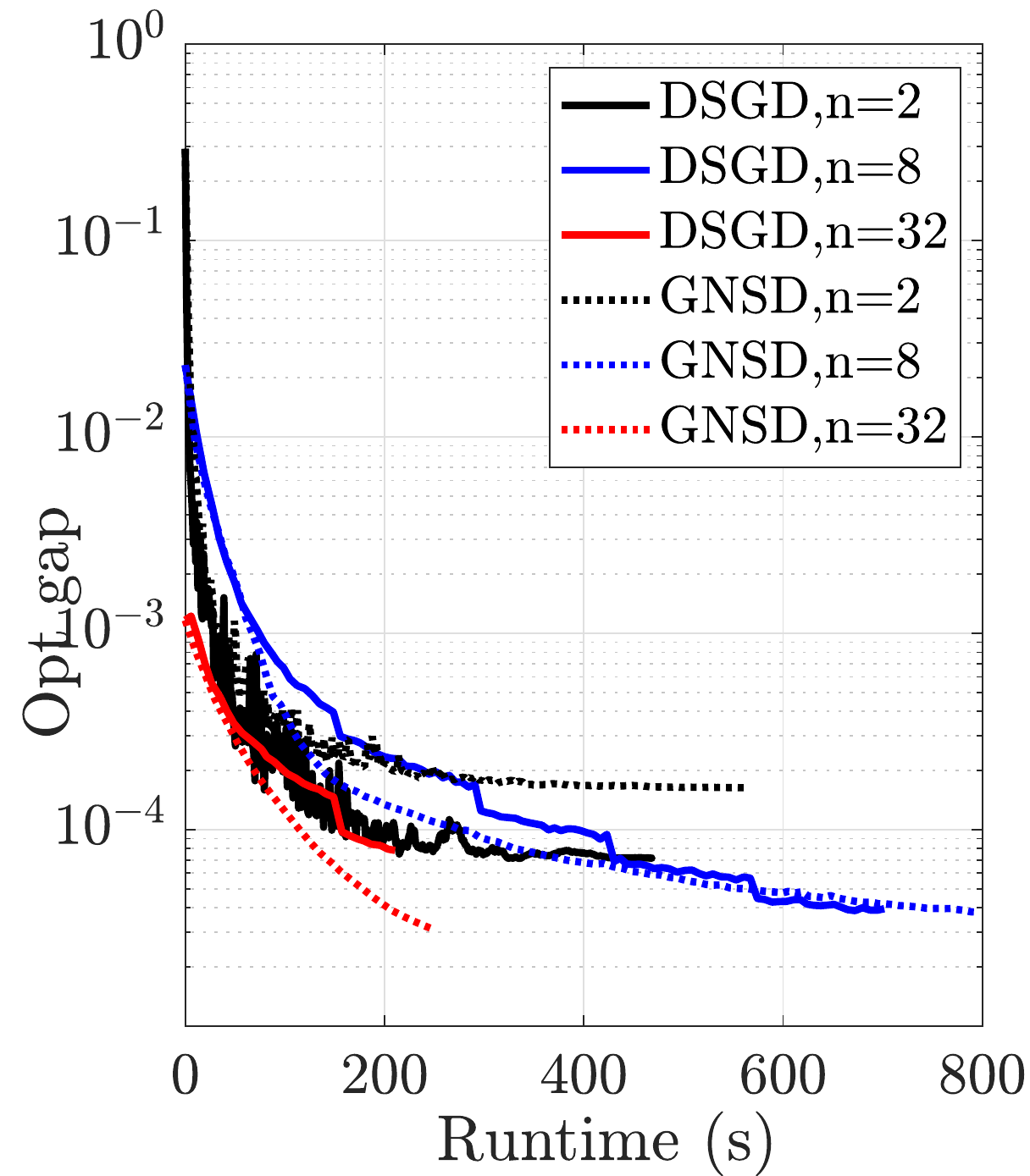}~
	\includegraphics[width=.28\linewidth]{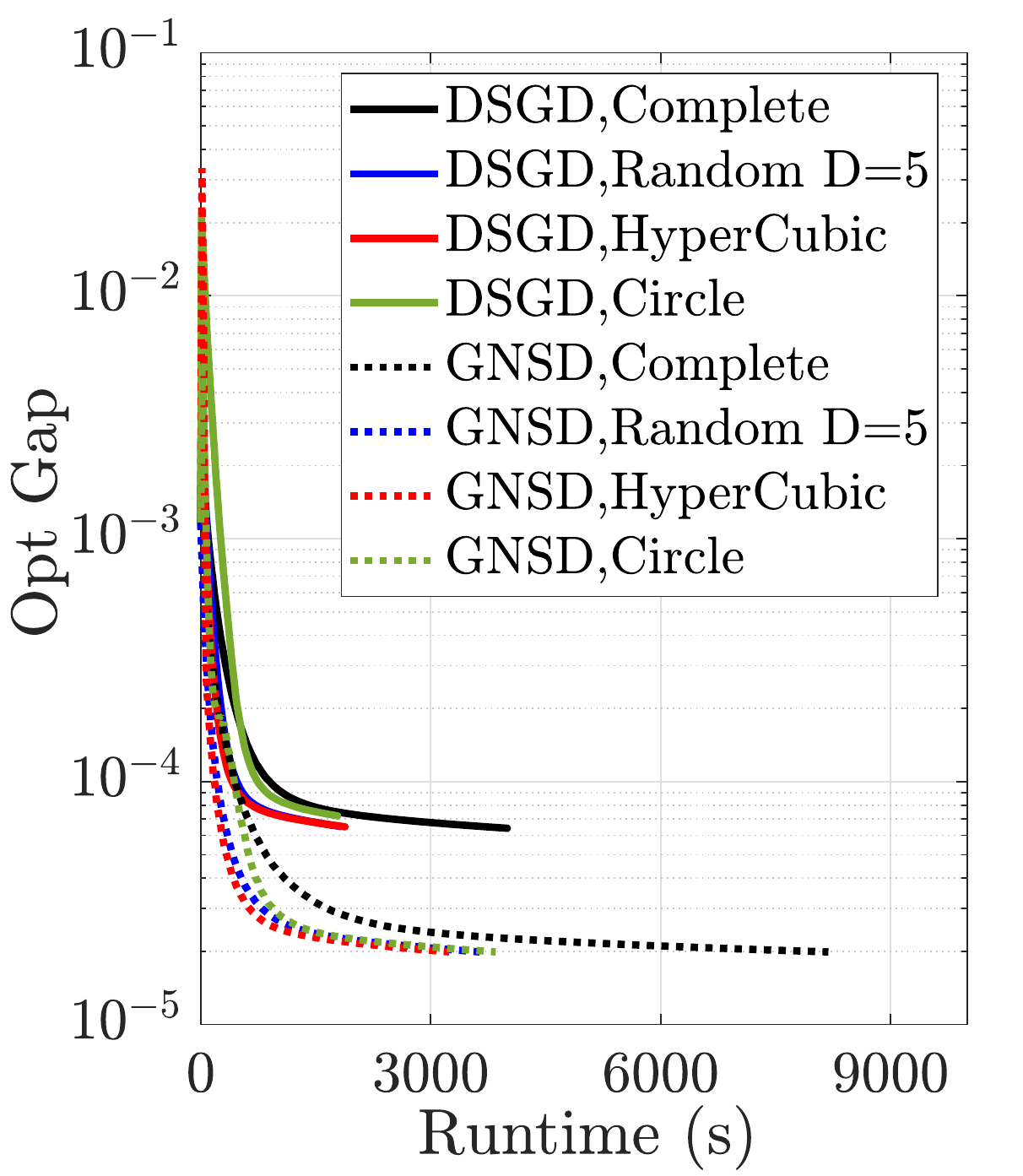}\vspace{-.3cm}
	\caption{\footnotesize Runtime comparison of streaming algorithms: (Left) on MSI with $n=1, 2, 8, 32$ agents,  batchsize $m=128$ for all algorithms, terminated in 450 epochs; (Middle) on AWS with $n = 2, 8, 32$ agents,  batchsize $m=128$, terminated in 128 epochs; (Right) on MSI with different types of graph topologies with $n=32$ agents, batchsize $m=128$, terminated in 256 epochs.}\vspace{-0.5cm}
	\label{fig:agent_num_a}
\end{figure}

{\bf Network Scalability.} We analyze how the network size $n$ affects the overall convergence speed. Intuitively, if the communication cost is relatively cheaper than that of computation, then it is beneficial to use a larger network and involve more agents to share the computational burden. %tj increasing the netowork size will benifit more than the case where communication is slower. 
In Fig.~\ref{fig:agent_num_a} (left) \& (middle), we see that the runtime performance of DSGD/GNSD algorithms on AWS significantly improves as the number of nodes increases (from $n=8$ to $n=32$); 
while there is no significant improvement for the experiments on MSI. This confirms our intuition since AWS has a high speed communication network.  {Besides, one can observe in the left figure the  benefit of distributed learning ($n>1$) over the centralized scheme ($n=1$), where DSGD with multiple agents can reach a smaller optimality gap}.
{On both platforms, we observe that GNSD achieves even smaller optimality gap compared with DSGD, but requires more time to complete the given number of epochs. This is reasonable since as discussed in Sec. \ref{sub:streaming}, DSGD requires one round of communication per evaluation of DO, while GNSD requires two.}

{\bf Graph Topology.} Another key parameter with a significant impact on the algorithm performance is the graph topology.  %When the  network size is the same, different the difference between network topologies will result in different performance of the algorithm. 
It is important to note that, although   theoretical analysis indicates that well-connected graphs [which have large $\xi(\Lb_G)$] has a faster convergence rate, in practice  factors such as the maximum degree of agents also matter.   %when we consider the communication in practice, sparse graphs will have shorter runtime due to shorter communication time per iteration as the number of neighbors is less in sparse graphs. 
In  Fig.~\ref{fig:agent_num_a} (right), we compare the runtime with $n=32$ agents on different types of topology -- including a complete graph, a random regular graph with degree 5, a hypercube graph, and a circle graph. We observe that well-connected {\it sparse} graphs (e.g., random regular, hybercube) are preferred, since there are less communication overheads compared with dense graphs (e.g., complete graph) and poorly-connected graphs (e.g., circle graph).

\begin{figure}[t]
	\centering
	\includegraphics[width=.37\linewidth]{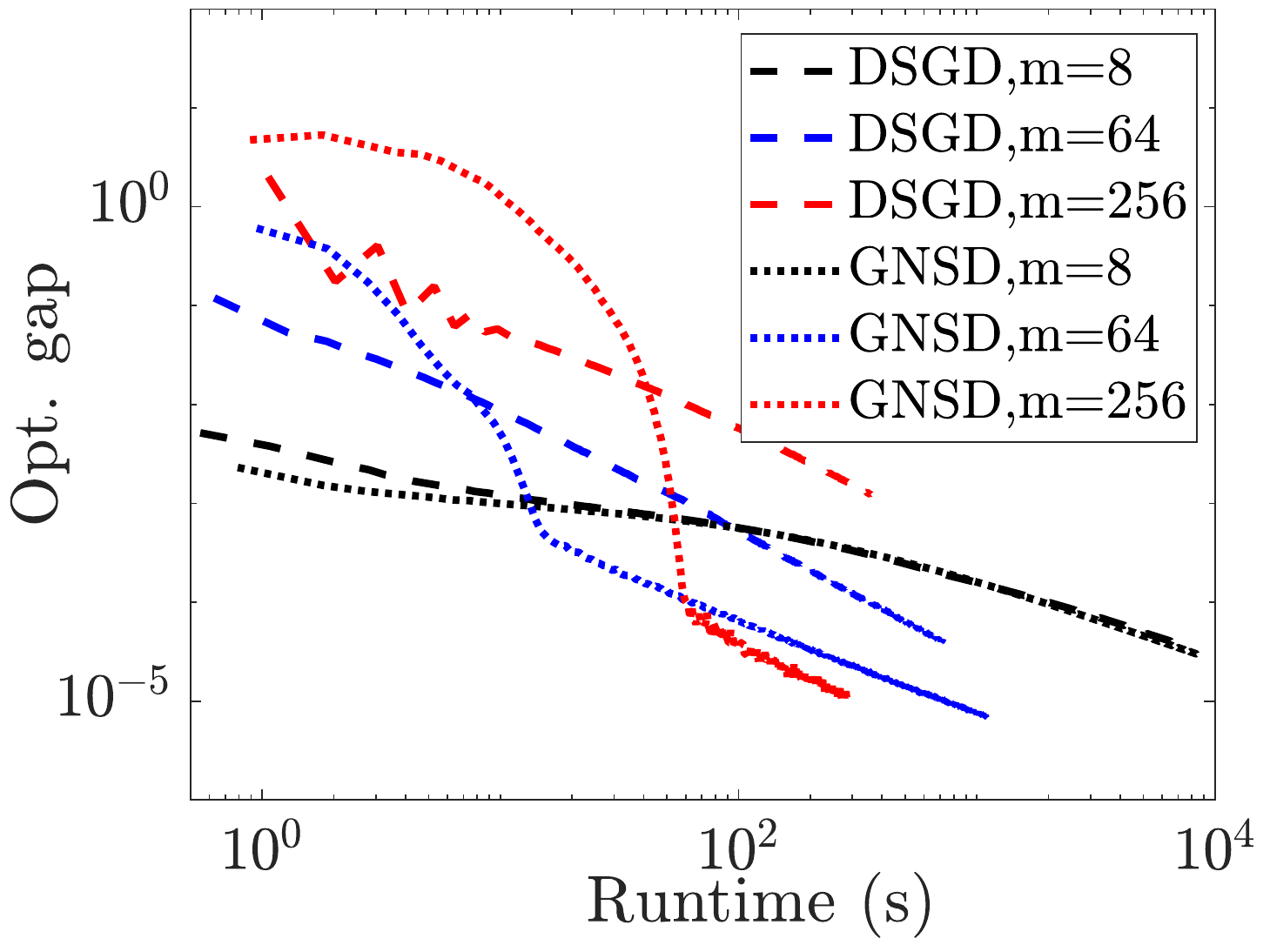}\quad\quad\includegraphics[width=.375\linewidth]{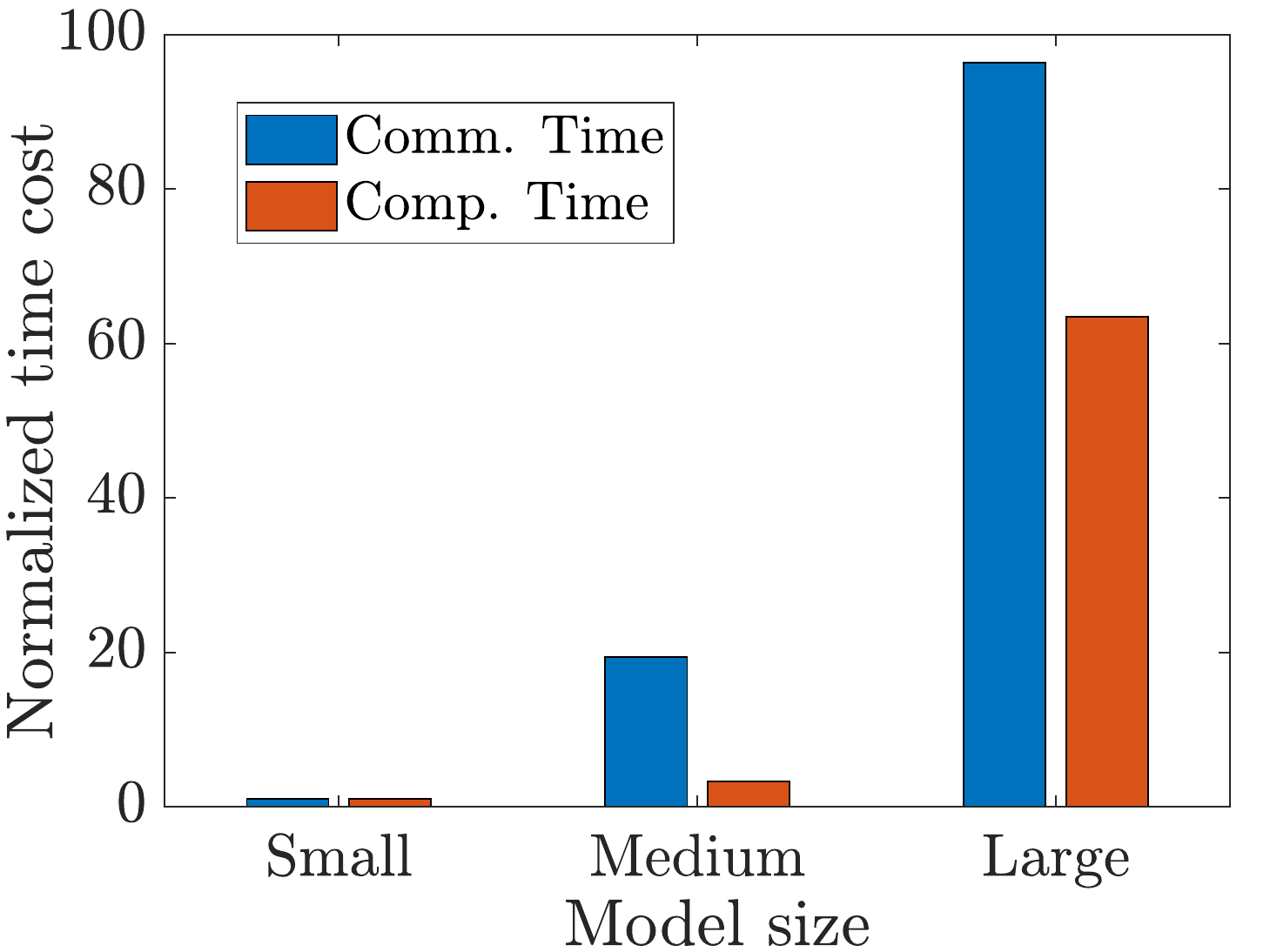}\vspace{-.3cm}
	\caption{\footnotesize (Left) Runtime comparison of mini-batch size $m=8, 64, 256$ on MSI, terminated after 256 epochs. Data is heterogeneous, where each node is assigned exclusive classes; (Right) Normalized computation and communication cost (normalized to the small model) for different model sizes on MSI with $n=32$ agents.}\vspace{-1cm}
	\label{fig:model_size}
\end{figure}

{\bf Mini-Batch Size.}
The choice of mini-batch size $m$ is another important parameter. While it speeds up the convergence with a large mini-batch size, it can be computationally expensive and requires extensive memory.  
We examine the tradeoff with the mini-batch size in Fig.~\ref{fig:model_size} (left), where the experiments are run on the MSI cluster. As seen, increasing the batch size improves the GNSD algorithm more significantly than DSGD. Further, in the lower part of Fig.~\ref{fig:matlab_alg} (right), we provide the normalized per-iteration computation and communication time with different mini-batch sizes. Notice that for DSGD, it takes $1.09$ and $1.36$ times of \emph{computation} time with a mini-batch size of $m=64$ and $m=256$, compared to the baseline setting with $m=8$. 
%\footnote{Such nonlinear increase is because the computation cost consists of a constant overhead and a data processing time proportional to the mini-batch size, therefore it is beneficial to use large mini-batch size.}.
A larger mini-batch size seems to be more  efficient.  %{\red[het-data?]}

%{\bf Heterogeneous Data.} 

{{\bf Heterogeneous Data.} We illustrate the effect of heterogeneous data on different algorithms by again using Fig. \ref{fig:model_size}. In this experiment, we divide the data according to  their labels, and assign each agent exclusively with two classes. We can see that the performance of DSGD becomes significantly worse compared with GNSD, especially when the batch size becomes larger (in which case the variance caused by sampling becomes smaller, hence the effect of heterogeneous data is more pronounced). This observation corroborates the theoretical results in Sec. \ref{sub:streaming}, that GNSD does not require any assumption on the distribution of the data, while DSGD does.} %entire class of data to one agent.  % agents are assigned data points with different labels. 

{\bf Model Size.}
Intuitively, small model may benefit from distributed algorithm due to the small amount of information exchange required, especially on systems where communication is slower than computation. As shown in Fig~\ref{fig:model_size} (right), we compare 3 neural networks -- a small network (2-layer fully connected neural network, with $8\times10^3$ parameters), a medium network (LeNet-5 with 2 convolutional layers and 3 fully connected layers, with $6 \times 10^4$ parameters), a large network ({Keras example for MNIST with 4 convolutional layers} and 3 fully connected layers, with $4.07 \times 10^5$ parameters), run on the MSI cluster with DSGD. As model size increases, the growth of communication cost outweighs   the computation cost.

%, which indicates that for large-sized model, more time will be spent on the communication. On the other hand, if the computation complexity of a model is superlinear to the model size (i.e, RNN model, Quantized model), then larger network may take more advantages from the growing network size. 
%{\red[can we have one example that demonstrate this? maybe just run on Aamzon? This is important, seems that we cannot leave this as a speculation.]}%{\blue[Tensorflow has some bug dealing with RNN network or network that requires integer input, so we can only leave this as a speculation or delete it.]}

\paragraph{Other Related Issues}
Another active research direction is on improving communication efficiency in distributed algorithms. 
Taking the DSGD as an example, a possible idea is to perform SGD updates locally for multiple times (say $I$) at an agent before exchanging the parameters with neighbors. Using this scheme, \cite{yu2019linear} shows that with $I = \Theta(1/\epsilon)$, the distributed algorithm run on a \emph{star graph} topology requires only ${\cal O}(1/\epsilon)$ [\resp ${\cal O}(1/\epsilon^{\frac{3}{2}})$] message exchanges for homogeneous (\resp heterogeneous) dataset to find an $\epsilon$-stationary solution to \eqref{eq:opt}.
Alternatively, \cite{Aragues12} proposes to skip unnecessary communication steps when the deviation of local variables is small.
{Lastly, to reduce the time cost on synchronizing over agents and to make distributed learning less vulnerable to straggling agents, 
	there are works that allow for asynchronous communication; see
	\cite{assran19,DBLP:conf/icml/LianZZL18} for example.}

\vspace{-0.3cm}
\section{Conclusions \& Open Problems} \label{sub:discussion}\vspace{-.2cm}
{This paper provides a selected review on recent developments of non-convex, distributed learning algorithms.}
{We show the interplay between \emph{problem, data} and \emph{computation, communication}, leading to different algorithms}. We also compare the algorithms using numerical experiments on computer clusters, showing their practical potentials. {Below we list a few directions for future research.}

\paragraph{Dynamical Data} 
Beyond batch and streaming data, an open problem is to develop distributed algorithms for \emph{dynamical} data. 
%A typical example where it applies is the policy optimization in reinforcement learning. Here, the scenario can be interpreted as the agent {actively} updating the iterates while \emph{interacting} with the environment and other agents. 
We consider a DO which takes the same form as the first equation of \eqref{eq:stream}, but the data samples $\{ \State_{i,\ell}^{t+1} \}_{\ell=1}^{M_\ell}$ are drawn instead from a \emph{parameterized} distribution $\pi_i( \bm{\cdot}; \param^t)$.
The new data model corresponds to a \emph{dynamic} data acquisition process controlled by the iterates. The output of this DO will be used by the algorithm to compute the next iterate.
For example, this is relevant to policy optimization where $\param^t$ is the joint policy exercised by the agents, and the data acquired are state/action pairs generated through interactions with the environment (therefore dependent on the current policy $\param^t$); the state/action pairs will then be used to compute the policy gradient for updating $\param^{t+1}$. 

% drawing data samples from a \emph{parameterized} distribution $\pi_i( \bm{\cdot}; \param^t)$. It can be seen that this models a data acquisition process dependent on the agents-controlled iterates $\param^t$. 
% the stochastic cost function is  defined as $f_i( \prm_i ) = \EE_{ \xi_i \sim \pi_i (\bm{\cdot} ; \prm_i ) } \big[ F_i ( \prm_i ; \xi_i ) \big]$.

Distributed algorithms based on the dynamic DO is challenging to analyze as computation, communication, and data acquisition have to be jointly considered. To the best of our knowledge, such setting has only been recently studied for a centralized algorithm in \cite{karimi2019non}. In a distributed setting, progresses have been made in multi-agent reinforcement learning, e.g., \cite{zhang2018fully} applied a linear function approximation to simplify the non-convex learning problem as a convex one. Nevertheless, a truly distributed, non-convex algorithm with a dynamic DO has neither been proposed nor analyzed. 
{ Another challenging dynamic scenario is under the \emph{online} setting, where no statistical assumption is imposed on the DO output. 
	%, and therefore the interest mainly on the worst-case gap between the online algorithm and the offline algorithm known as the dynamic regret. 
	However, most of the developments are still restricted to convex problems; see \cite{shahrampour2017distributed}.}
%\cite{mokhtari2016online,shahrampour2017distributed}.} %\mhcomment{if no space remove?}
%{\blue Tsung-Hui: can we say online algorithms deal with some kind of "dynamic" data?} \mhcomment{ Requested reference\cite{shahrampour2017distributed}}

%This setting is the most challenging among all, where the data is acquired \emph{dynamically}.   To set this up, consider a general stochastic cost model with $f_i( \prm_i ) = \EE_{ \xi_i \sim \pi_i (\bm{\cdot} ; \prm_i ) } \big[ F_i ( \prm_i ; \xi_i ) \big]$,
%where the probability distribution $\pi_i (\bm{\cdot} ; \prm_i )$ is parameterized by $\prm_i$. 
%Like in the streaming data setting, the DO takes the same stochastic form as in the  first equation of \eqref{eq:stream}. However, the data is not i.i.d.~and the acquisition process is \emph{controlled} by the iterates $\prm_i^{t}$. To model this, define $\param = (\prm_1,...,\prm_n)$ and we follow a Markovian model with a transition matrix $P(\bm{\cdot}, \bm{\cdot}; \prm )$, such that the unique stationary distribution is the desired $\pi( \bm{\cdot} ; \prm )$. The random samples in the DO  \eqref{eq:stream} is drawn from the state-controlled Markov chain:
%\beq \label{eq:dynamic}
%\State^{t,\ell+1} \sim P( \State^{t,\ell}, {\bm{\cdot}} \!~; \param^{t} ),~\ell=1,...,m_t-1,~\text{where}~~\State^{t,1} \sim P(\State^{t-1,m_{t-1}} , \bm{\cdot} \!~; \param^{t} ).
%\eeq
%Unlike in the streaming data setting, the DO produces \emph{biased} gradient estimates $\EE \big[ {\sf DO}_i ( \prm_i^{t} ) \big] \neq \grd f_i( \prm_i^{t} )$.

\paragraph{Distributed Feature}
%Problem \eqref{eq:opt} and \eqref{eq:consensus} assumed that the local data at the agents share the same set of features.
{In many applications, leveraging additional features from another domain or party can further improve the inference performance.
	%; for example, all $\xb_i^{t,\ell}$ in Example 1 are bank transaction records of clients for all $i,t,\ell$. 
	However, data with these features may be private records and cannot be shared. }
This imposes a challenging question of how to enable the agents that own different sets of features to collaborate on the learning task; see \cite{distributedfeature2019,chang14distributed}. 

%Taking \eqref{eq:logist loss} as the example, this problem corresponds to the one in which the input to the 2nd layer of the neural network is $\Wb^{(1)}_1\xb_1^{t,\ell}+\cdots+\Wb^{(1)}_n\xb_n^{t,\ell}$, where $\Wb^{(1)}=[\Wb^{(1)}_1,\ldots, \Wb^{(1)}_n]$ contains the 1st layer neural network parameters. {\red[what's the challenges.]}

\paragraph{Federated \& Robust Learning}
{To improve user privacy, federated learning (FL) \cite{Federatedlearning2016} is proposed for distributed learning in edge networks. 
	Unlike traditional distributed learning, FL emphasizes on the ability to deal with unbalanced data and poorly connected users.
	Security is another concern for FL and algorithms that are resilient to adversary attacks or model poisoning are crucial \cite{yang2019byrdie} for example.}

\paragraph{Distributed Learning with Statistical Guarantees}
The algorithms surveyed in this work aim at computing high-quality solutions, in the sense that optimization based conditions such as \eqref{eq:stationarity} are satisfied. It is also interesting to investigate whether these algorithms can achieve strong statistical guarantees for specific machine learning problems such as non-convex M-estimation \cite{Loh15}, so that  ground truth parameters can also be recovered. %We expect that for specific machine learning problems such as non-convex M-estimation \cite{Loh15}. %by properly leverage promising results obtained in the centralized case %ome interesting recent work can be found in 

\vspace{-1cm}
\section{Acknowledgement}
The authors would like to thank the anonymous reviewers, and Dr. Gesualdo Scutari for helpful comments that significantly improved the quality of the paper. 

\vspace{-0.4cm}

%	
%	It aims to handle training datasets that i) are distributed across a large number of mobile devices ($n\gg M_i$), ii) have non-i.i.d. distributions (since local data are far from enough to be representative for the whole data distribution), and iii) are unbalanced in the sense that the devices have different amounts of data samples. Besides, the mobile devices are frequently offline and poorly connected. 
%While several %communication-efficient 
%algorithms 
%\cite{Federatedlearning2017} have been proposed, the current focus of federated learning is still on the star network topology where one server coordinates massive mobile devices for learning a centralized model. %Furthermore, there is an increasing trend to study distributed algorithms that are resilient to adversary attacks \cite{yang2019byrdie}.
%
%Apart from communication efficiency, which is the primary concern of current studies, there is an increasing trend to strengthen the security of distributed learning by considering algorithms that are resilient to adversary attacks or model poisoning; see \cite{chen2017distributed,yang2019byrdie,bhagoji2018analyzing} for example.

%	Communication-Computation Tradeoff

	{\small \bibliographystyle{IEEEtran}
	\bibliography{combined_ref}}

\begin{thebibliography}{10}
\providecommand{\url}[1]{#1}
\csname url@samestyle\endcsname
\providecommand{\newblock}{\relax}
\providecommand{\bibinfo}[2]{#2}
\providecommand{\BIBentrySTDinterwordspacing}{\spaceskip=0pt\relax}
\providecommand{\BIBentryALTinterwordstretchfactor}{4}
\providecommand{\BIBentryALTinterwordspacing}{\spaceskip=\fontdimen2\font plus
\BIBentryALTinterwordstretchfactor\fontdimen3\font minus
  \fontdimen4\font\relax}
\providecommand{\BIBforeignlanguage}[2]{{%
\expandafter\ifx\csname l@#1\endcsname\relax
\typeout{** WARNING: IEEEtran.bst: No hyphenation pattern has been}%
\typeout{** loaded for the language `#1'. Using the pattern for}%
\typeout{** the default language instead.}%
\else
\language=\csname l@#1\endcsname
\fi
#2}}
\providecommand{\BIBdecl}{\relax}
\BIBdecl

\bibitem{Wen2017}
W.~Wen, C.~Xu, F.~Yan, C.~Wu, Y.~Wang, Y.~Chen, and H.~Li, ``Terngrad: Ternary
  gradients to reduce communication in distributed deep learning,'' in
  \emph{Advances in neural information processing systems}, 2017, pp.
  1509--1519.

\bibitem{Daily2018}
J.~Daily, A.~Vishnu, C.~Siegel, T.~Warfel, and V.~Amatya, ``{GossipGraD}:
  Scalable deep learning using gossip communication based asynchronous gradient
  descent,'' preprint, available at arXiv:1803.05880.

\bibitem{Jiang2017}
Z.~Jiang, A.~Balu, C.~Hegde, and S.~Sarkar, ``Collaborative deep learning in
  fixed topology networks,'' in \emph{Advances in Neural Information Processing
  Systems}, 2017.

\bibitem{nedic2010coop}
A.~Nedic and A.~Ozdaglar, ``Cooperative distributed multi-agent optimization,''
  in \emph{Convex Optimization in Signal Processing and Communications}.\hskip
  1em plus 0.5em minus 0.4em\relax Cambridge University Press, 2010.

\bibitem{sayed2013diffusion}
A.~H. Sayed, S.-Y. Tu, J.~Chen, X.~Zhao, and Z.~J. Towfic, ``Diffusion
  strategies for adaptation and learning over networks: an examination of
  distributed strategies and network behavior,'' \emph{IEEE Signal Process.
  Mag.}, vol.~30, no.~3, pp. 155--171, 2013.

\bibitem{cevher2014convex}
V.~Cevher, S.~Becker, and M.~Schmidt, ``Convex optimization for big data:
  Scalable, randomized, and parallel algorithms for big data analytics,''
  \emph{IEEE Signal Processing Magazine}, vol.~31, no.~5, pp. 32--43, 2014.

\bibitem{Murty1987}
\BIBentryALTinterwordspacing
K.~G. Murty and S.~N. Kabadi, ``Some {NP}-complete problems in quadratic and
  nonlinear programming,'' \emph{Mathematical Programming}, vol.~39, no.~2, pp.
  117--129, Jun 1987. [Online]. Available:
  \url{http://dx.doi.org/10.1007/BF02592948}
\BIBentrySTDinterwordspacing

\bibitem{hong2018gradient}
M.~Hong, M.~Razaviyayn, and J.~Lee, ``Gradient primal-dual algorithm converges
  to second-order stationary solution for nonconvex distributed optimization
  over networks,'' in \emph{ICML}, 2018, pp. 2014--2023.

\bibitem{daneshmand2018second-arxiv}
A.~Daneshmand, G.~Scutari, and V.~Kungurtsev, ``Second-order guarantees of
  distributed gradient algorithms,'' \emph{arXiv preprint arXiv:1809.08694},
  2018.

\bibitem{Swenson19annealing}
B.~Swenson, S.~Kar, H.~V. Poor, and J.~M.~F. Moura, ``Annealing for distributed
  global optimization,'' \emph{arXiv Preprint}, 2019, arXiv:1903.07258.

\bibitem{vlaski2019distributed}
S.~Vlaski and A.~H. Sayed, ``Distributed learning in non-convex
  environments--part {I}: Agreement at a linear rate,'' \emph{arXiv preprint
  arXiv:1907.01848}, 2019.

\bibitem{vlaski2019distributedb}
------, ``Distributed learning in non-convex environments--part {II}:
  Polynomial escape from saddle-points,'' \emph{arXiv preprint
  arXiv:1907.01849}, 2019.

\bibitem{Nedic09subgradient}
A.~Nedi{\'c} and A.~Ozdaglar, ``Distributed subgradient methods for multi-agent
  optimization,'' \emph{IEEE Transactions on Automatic Control}, vol.~54,
  no.~1, pp. 48--61, 2009.

\bibitem{boyd2004fastest}
S.~Boyd, P.~Diaconis, and L.~Xiao, ``Fastest mixing markov chain on a graph,''
  \emph{SIAM review}, vol.~46, no.~4, pp. 667--689, 2004.

\bibitem{Hong17ICML}
M.~Hong, D.~Hajinezhad, and M.-M. Zhao, ``{Prox-PDA}: The proximal primal-dual
  algorithm for fast distributed nonconvex optimization and learning over
  networks,'' in \emph{ICML}, 2017.

\bibitem{chang14distributed}
T.-H. Chang, M.~Hong, and X.~Wang, ``Multi-agent distributed optimization via
  inexact consensus {ADMM},'' \emph{IEEE Transactions on Signal Processing},
  vol.~63, no.~2, pp. 482--497, Jan 2015.

\bibitem{Zeng19distributedGD}
J.~Zeng and W.~Yin, ``On nonconvex decentralized gradient descent,'' \emph{IEEE
  Transactions on Signal Processing}, vol.~66, no.~11, pp. 2834--2848, June
  2018.

\bibitem{shi14extra}
W.~Shi, Q.~Ling, G.~Wu, and W.~Yin, ``{EXTRA}: An exact first-order algorithm
  for decentralized consensus optimization,'' \emph{SIAM Journal on
  Optimization}, vol.~25, no.~2, pp. 944--966, 2014.

\bibitem{sun18optimal}
H.~Sun and M.~Hong, ``Distributed non-convex first-order optimization and
  information processing: Lower complexity bounds and rate optimal
  algorithms,'' \emph{IEEE Transactions on Signal processing}, July 2019,
  accepted for publication.

\bibitem{scutari2019distributed}
G.~Scutari and Y.~Sun, ``Distributed nonconvex constrained optimization over
  time-varying digraphs,'' \emph{Mathematical Programming}, vol. 176, no. 1-2,
  pp. 497--544, 2019.

\bibitem{nedic2017achieving}
A.~Nedic, A.~Olshevsky, and W.~Shi, ``Achieving geometric convergence for
  distributed optimization over time-varying graphs,'' \emph{SIAM Journal on
  Optimization}, vol.~27, no.~4, pp. 2597--2633, 2017.

\bibitem{tsitsiklis86}
J.~Tsitsiklis, D.~P. Bertsekas, and M.~Athans, ``Distributed asynchronous
  deterministic and stochastic gradient optimization algorithms,'' \emph{IEEE
  Transactions on Automated Control}, vol.~31, pp. 803--812, 1986.

\bibitem{cattivelli2009diffusion}
F.~S. Cattivelli and A.~H. Sayed, ``Diffusion {LMS} strategies for distributed
  estimation,'' \emph{IEEE Transactions on Signal Processing}, vol.~58, no.~3,
  pp. 1035--1048, 2009.

\bibitem{kar2012distributed}
S.~Kar, J.~M. Moura, and K.~Ramanan, ``Distributed parameter estimation in
  sensor networks: Nonlinear observation models and imperfect communication,''
  \emph{IEEE Transactions on Information Theory}, vol.~58, no.~6, pp.
  3575--3605, 2012.

\bibitem{Lian17}
X.~Lian, C.~Zhang, H.~Zhang, C.-J. Hsieh, W.~Zhang, and J.~Liu, ``Can
  decentralized algorithms outperform centralized algorithms? a case study for
  decentralized parallel stochastic gradient descent,'' in \emph{NeurIPS},
  2017, pp. 5330--5340.

\bibitem{Tang18}
H.~Tang, X.~Lian, M.~Yan, C.~Zhang, and J.~Liu, ``{D$^2$}: Decentralized
  training over decentralized data,'' in \emph{Proc. of the 35th International
  Conference on Machine Learning}, 10--15 Jul. 2018, pp. 4848--4856.

\bibitem{zhang2019decentralized}
J.~Zhang and K.~You, ``Decentralized stochastic gradient tracking for empirical
  risk minimization,'' \emph{arXiv preprint arXiv:1909.02712}, 2019.

\bibitem{luho19}
S.~{Lu}, X.~{Zhang}, H.~{Sun}, and M.~{Hong}, ``{GNSD}: {A} gradient-tracking
  based nonconvex stochastic algorithm for decentralized optimization,'' in
  \emph{Proc. of IEEE Data Science Workshop (DSW)}, Jun. 2019, pp. 315--321.

\bibitem{daneshmand2018decentralized}
A.~Daneshmand, Y.~Sun, G.~Scutari, F.~Facchinei, and B.~M. Sadler,
  ``Decentralized dictionary learning over time-varying digraphs,''
  \emph{Journal of Machine Learning Research}, 2019.

\bibitem{assran19}
M.~Assran, N.~Loizou, N.~Ballas, and M.~Rabbat, ``Stochastic gradient push for
  distributed deep learning,'' in \emph{Proc. of International Conference on
  Machine Learning}, 2019, pp. 344--353.

\bibitem{yu2019linear}
H.~Yu, R.~Jin, and S.~Yang, ``On the linear speedup analysis of communication
  efficient momentum sgd for distributed non-convex optimization,'' in
  \emph{International Conference on Machine Learning}, 2019, pp. 7184--7193.

\bibitem{Aragues12}
R.~Aragues, G.~Shi, D.~V. Dimarogonas, C.~Sagues, and K.~H. Johansson,
  ``Distributed algebraic connectivity estimation for adaptive event-triggered
  consensus,'' in \emph{2012 American Control Conference (ACC)}, June 2012, pp.
  32--37.

\bibitem{DBLP:conf/icml/LianZZL18}
X.~Lian, W.~Zhang, C.~Zhang, and J.~Liu, ``Asynchronous decentralized parallel
  stochastic gradient descent,'' in \emph{Proc. of ICML}, July 10-15, 2018, pp.
  3049--3058.

\bibitem{karimi2019non}
B.~Karimi, B.~Miasojedow, E.~Moulines, and H.-T. Wai, ``Non-asymptotic analysis
  of biased stochastic approximation scheme,'' in \emph{Conference on Learning
  Theory}, 2019.

\bibitem{zhang2018fully}
K.~Zhang, Z.~Yang, H.~Liu, T.~Zhang, and T.~Ba{\c{s}}ar, ``Fully decentralized
  multi-agent reinforcement learning with networked agents,'' in
  \emph{International Conference on Machine Learning}, 2018, pp. 9340--9371.

\bibitem{shahrampour2017distributed}
S.~Shahrampour and A.~Jadbabaie, ``Distributed online optimization in dynamic
  environments using mirror descent,'' \emph{IEEE Transactions on Automatic
  Control}, vol.~63, no.~3, pp. 714--725, 2017.

\bibitem{distributedfeature2019}
Y.~Hu, D.~Niu, J.~Yang, and S.~Zhou, ``{FDML}: {A} collaborative machine
  learning framework for distributed features,'' in \emph{Proc. ACM
  International Conference on Knowledge Discovery \& Data Mining}, Aug. 2018,
  pp. 2232--2240.

\bibitem{Federatedlearning2016}
J.~Konecny, H.~B. McMahan, and D.~Ramage, ``Federated optimization:
  {D}istributed optimization beyond the datacenter,'' in \emph{Proc. of
  Optimization for Machine Learning}, 2015, pp. 1--5.

\bibitem{yang2019byrdie}
Z.~Yang and W.~U. Bajwa, ``Byrdie: Byzantine-resilient distributed coordinate
  descent for decentralized learning,'' \emph{IEEE Transactions on Signal and
  Information Processing over Networks}, 2019.

\bibitem{Loh15}
P.-L. Loh and M.~J. Wainwright, ``Regularized m-estimators with nonconvexity:
  Statistical and algorithmic theory for local optima,'' \emph{Journal of
  Machine Learning Research}, vol.~16, no.~19, pp. 559--616, 2015.

\end{thebibliography}
%	\bibliography{ref,ref2,refs_ml,ref_tiancong,opt_bib,ref_th}
	
	%-----------------------------------------------------------------------------
	%\vspace{0.4cm}
	
\end{document}